\pdfoutput=1
\documentclass[11pt]{article}

\usepackage{authblk}
\usepackage[final]{acl}

\usepackage{times}
\usepackage{latexsym}

\usepackage{physics,amsmath}
\usepackage{cleveref}
\usepackage{enumitem}
\usepackage{booktabs}

\usepackage{textcomp, xspace}
\usepackage{pbox}
\usepackage{array}
\usepackage{multirow}
\usepackage{xcolor}
\usepackage[normalem]{ulem}
\useunder{\uline}{\ul}{}
\usepackage{rotating}
\usepackage{array}
\usepackage{listings}
\usepackage{adjustbox}

\usepackage{arydshln}
\usepackage[most]{tcolorbox}
\usepackage{csquotes}
\usepackage{fvextra}

\usepackage[T1]{fontenc}
\usepackage[utf8]{inputenc}

\usepackage{microtype}

\usepackage{inconsolata}

\usepackage{graphicx}

\newcommand{\framework}{\textsc{PRIME}\xspace}
\newcommand{\data}{\textsc{CMV}\xspace}

\title{\framework: Large Language Model Personalization with \\Cognitive Dual-Memory and Personalized Thought Process}

\author[1]{\textbf{Xinliang Frederick Zhang}}
\author[2]{\textbf{Nick Beauchamp}}
\author[1]{\textbf{Lu Wang}}

\affil[1]{Computer Science and Engineering, University of Michigan, Ann Arbor, MI}
\affil[2]{Department of Political Science, Northeastern University, Boston, MA}

\affil[ ]{$^1$\{\texttt{xlfzhang,wangluxy\}@umich.edu}, $^2$\texttt{n.beauchamp@northeastern.edu}}

\begin{document}
\maketitle

\begin{abstract}
Large language model (LLM) personalization aims to align model outputs with individuals' unique preferences and opinions. 
While recent efforts have implemented various personalization methods, a unified theoretical framework that can systematically understand the drivers of effective personalization is still lacking. 
In this work, we integrate the well-established cognitive dual-memory model into LLM personalization, by mirroring \textit{episodic memory} to historical user engagements and \textit{semantic memory} to long-term, evolving user beliefs. 
Specifically, we systematically investigate memory instantiations and introduce a unified framework, \framework, using episodic and semantic memory mechanisms. 
We further augment \framework with a novel \textit{personalized thinking} capability inspired by the slow thinking strategy. 
Moreover, recognizing the absence of suitable benchmarks, we introduce a dataset using Change My View (\data) from Reddit\footnote{Our project page can be found at \url{http://github.com/launchnlp/LM_Personalization}.},
specifically designed to evaluate long-context personalization. 
Extensive experiments validate \framework's effectiveness across both long- and short-context scenarios. 
Further analysis confirms that \framework effectively captures dynamic personalization beyond mere popularity biases.
\end{abstract}

\section{Introduction}

Personalization~\citep{DBLP:journals/datamine/SchaferKR01, DBLP:conf/intetain/BerkovskyKR05} aims to tailor model outputs to individual users' needs, preferences and beliefs, moving beyond generic responses~\citep{zhang-etal-2018-personalizing, DBLP:conf/www/HuangFDWASY22, tseng-etal-2024-two}. 
While large language models (LLMs) excel at diverse NLP tasks, users' demand for personalized LLMs that reflect their unique histories and preferences has grown~\citep{salemi-etal-2024-lamp, liu2025survey}.
 For instance, we have seen personalization adopted into commercial applications, such as OpenAI's customizable GPTs,\footnote{\url{https://openai.com/index/introducing-gpts/}} which are essential for building trust and reducing interaction friction~\citep{Castells15}. 
In this work, we formally define a \textit{personalized LLM} as one adapted to align with the individual preferences, characteristics, and beliefs, by utilizing user-specific attributes, past engagements, and context the user was exposed to~\citep{Zhang2024PersonalizationOL}.
Various techniques have been explored for LLM personalization, including prompt engineering~\citep{DBLP:journals/corr/abs-2306-11114, DBLP:journals/corr/abs-2305-06474}, retrieval-augmented generation~\citep{salemi-etal-2024-lamp, mysore-etal-2024-pearl}, efficient fine-tuning~\citep{tan-etal-2024-democratizing, DBLP:journals/corr/abs-2404-03565}, and reinforcement learning from human feedback~\citep{DBLP:journals/corr/abs-2402-05133}. Yet these piecemeal approaches lack a unified framework for systematically identifying what makes personalization effective. 
We posit that drawing inspiration from established cognitive models of human memory~\citep{ATKINSON196889} offers a principled way to understand and advance LLM personalization. 
Specifically, we propose a \textbf{dual-memory model}~\citep{tulving1972episodic, tulving1985many, schacter2009semantic} with \textit{episodic memory} (specific personal experiences) and \textit{semantic memory} (abstract knowledge and beliefs) that parallels existing LLM personalization techniques. 

\begin{figure*}[t]
    \centering
    \includegraphics[width=0.8\textwidth]{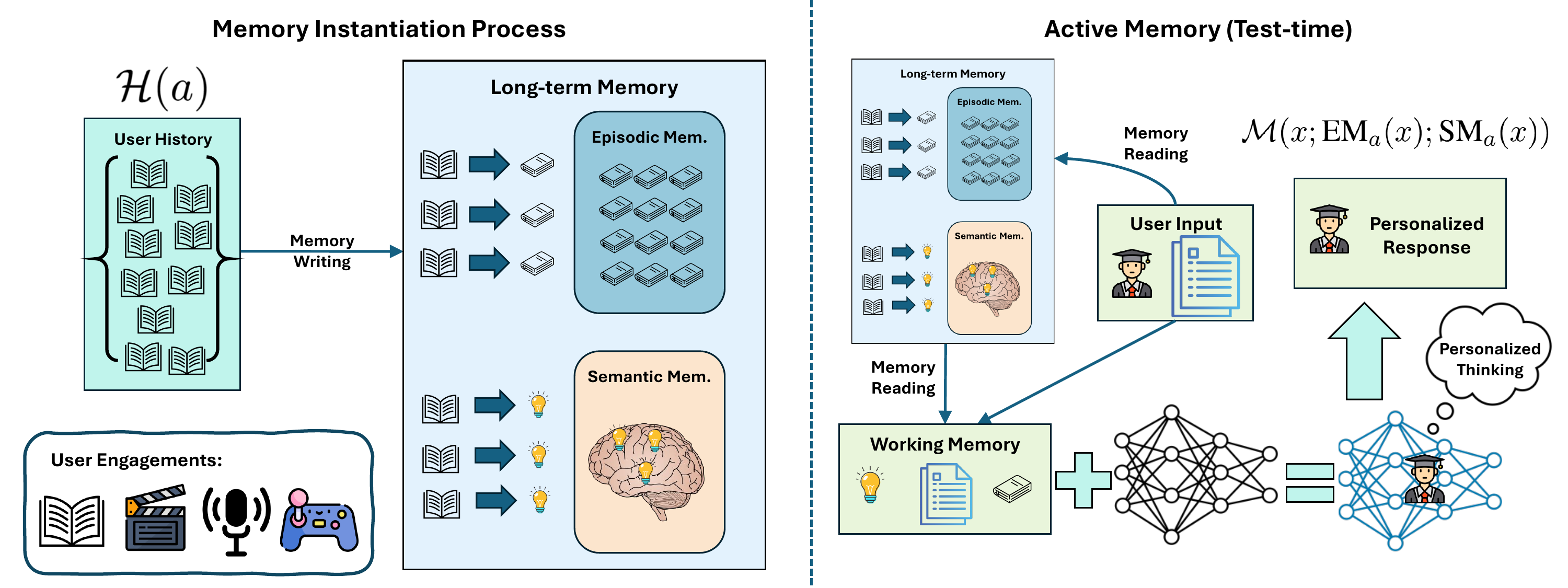}
    \vspace{-2mm}
    \caption{
    Overview of our unified framework, \framework, inspired by dual-memory model~\citep{tulving1972episodic}. \framework is further augmented with {personalized thinking}, yielding more accurate, user‐aligned responses.
    }
    \vspace{-4mm}
    \label{fig:framework}
\end{figure*}

Based on the cognitive model,
we begin by examining memory instantiations to understand their strengths and weaknesses. 
Next, we present a unified framework, dubbed \textbf{\framework} (\textbf{P}ersonalized \textbf{R}easoning with \textbf{I}ntegrated \textbf{ME}mory),  
to integrate both memory mechanisms in a principled manner (Figure~\ref{fig:framework}). 
Such integration facilitates a holistic understanding of user queries and histories, enabling the model to generate responses that are both contextually relevant and aligned with the user's long-term beliefs.
Furthermore, within \framework, we introduce the generation of chain-of-thoughts (CoTs) using \textbf{personalized thinking}, which draws on 
the slow thinking strategy~\citep{muennighoff2025s1, chen2025empirical}. 
Yet, we find that generic CoT reasoning can hinder performance on tasks that require personalized perspectives~\citep{guo2025deepseek}. 
In contrast, by adapting the self-distillation strategy~\citep{zhang2019your, pham2022revisiting, wang-etal-2023-self-instruct}, we unlock LLM's \textit{personalized thinking capability}. 
This ability guides the model to perform customized reasoning, yielding more accurate, user‐aligned responses and richer reasoning traces that reflect the user’s history and traits.

Meanwhile, benchmarking LLM personalization capabilities is hindered by a lack of suitable datasets~\citep{tseng-etal-2024-two}. Most datasets focus on short-context queries and surface‐level imitation~\citep[e.g., stylistic mimicry;][]{wu-etal-2020-mind, salemi-etal-2024-lamp}, neglecting genuine personalization---\textit{users’ latent beliefs and perspectives}---which requires modeling deeper, long‐term preferences and traits.
To this end, we introduce a novel dataset derived from the \textit{Change My View} (CMV) Reddit forum\footnote{\url{https://www.reddit.com/r/changemyview/}},  which comprises $133$ 
challenging evaluation posts by 41 active authors, along with their $7,514$ instances of historical engagements. 
CMV discussions feature \textit{extended dialogues} where participants seek to change the original poster’s (OP's) opinion on varied topics.
We cast the interactions into a ranking-based recommendation task, where the objective is to identify the response that effectively alters the OP's point of view, as acknowledged by the OP. 

% \begin{figure*}[t]
%     \centering
%     \vspace{-4mm}
%     \includegraphics[width=0.8\textwidth]{figures/framework.pdf}
%     \vspace{-2mm}
%     \caption{
%     Overview of our unified framework, \framework, inspired by dual-memory model~\citep{tulving1972episodic}. \framework is further augmented with {personalized thinking}, yielding more accurate, user‐aligned responses.
%     }
%     \label{fig:framework}
% \end{figure*}

We conduct extensive empirical experiments on our curated \data data 
and an existing LLM personalization benchmark---LaMP~\citep{salemi-etal-2024-lamp}. Results show that 1) semantic memory model behaves generally more robust than episodic memory model; 2) our proposed \framework is compatible with models of different families and sizes, yielding better results than competitive comparisons; 3) and \textit{personalized thinking} plays a pivotal role in improving personalization.
Further analysis also demonstrates that personalized thinking can be enabled in training-free settings, offering flexibility in handling users with limited history which is often framed as the ``cold-start'' challenge~\citep{DBLP:journals/corr/abs-2501-01945}. 
To assess how effective our models capture user‐specific characteristics, we inject other users’ histories and measure the resulting performance drop, confirming that our method captures dynamic personalization rather than bandwagon biases.

In summary, our contributions are threefold:
\begin{itemize}[leftmargin=2em,itemsep=0em,topsep=1pt,parsep=1pt,partopsep=1pt]
    \item We propose \framework, a cognitively inspired unified framework for LLM personalization, further augmented with {personalized thinking}.
    \item We introduce a challenging dataset, derived from the CMV forum, with nuanced user beliefs and preferences in long-context setting.
    \item Experiments showcase the effectiveness of \framework, as well as the pivotal role of personalized thinking.
\end{itemize}

\section{Related Work}

\subsection{LLM Personalization}
\label{sec:llm_personalization}
\paragraph{Methods for Personalization.}
Early personalization in NLP relied on explicit user models, i.e., a structured representation of user traits, to tailor system outputs~\citep{DBLP:conf/ercimdl/AmatoS99,DBLP:journals/corr/abs-2402-09660}. They rely on static demographic features (e.g., age, gender, location) and use hand-crafted rules to adapt outputs~\citep{DBLP:conf/chi/GouZY14, kim2013personality, DBLP:conf/recsys/GaoHBLLZ13}. Latent-factor techniques like matrix factorization~\citep{DBLP:journals/computer/KorenBV09, DBLP:journals/tkde/JiangCWZY14}
decompose the user-item interaction matrix into low‐dimensional embeddings. Moreover, the Transformer architecture~\citep{transformer} enables and advances learnable user embedding approaches~\citep{DBLP:conf/aaai/QiuWG021, deng-etal-2023-annotate}. 
However, these methods all overlook \textit{unstructured} user-written content and fail to generalize across tasks, yielding shallow, brittle personalization and underscoring the need for more robust methods.

With LLMs, three major paradigms have emerged: prompt engineering, retrieval-augmented generation, and training-based parameterization.
Prompt‐based approaches prepend user context, such as profile summaries~\citep{DBLP:journals/corr/abs-2310-20081} or past interactions~\citep{DBLP:journals/corr/abs-2304-10149,DBLP:journals/corr/abs-2306-11114,DBLP:journals/corr/abs-2305-06474}, to the model input, but this method is constrained by LLMs' context window sizes. 
An improved version relies on retrievers like BM25~\citep{DBLP:journals/ftir/RobertsonZ09} and FAISS~\citep{douze2024faiss} to fetch relevant user history, which is then included in the model input~\citep{madaan-etal-2022-memory, salemi-etal-2024-lamp, mysore-etal-2024-pearl}. However, noisy or irrelevant retrieval limits their ability to capture fine‐grained user preferences. 
To address these challenges, recent studies have proposed to parameterize the historical engagement through training, by learning embeddings~\citep{doddapaneni-etal-2024-user, DBLP:journals/corr/abs-2402-13598}, by fine-tuning light-weight adapters~\citep{tan-etal-2024-democratizing,DBLP:journals/corr/abs-2404-03565}, or by employing RLHF to align with individuals’ preferences~\citep{DBLP:conf/nips/ChristianoLBMLA17, DBLP:conf/nips/Ouyang0JAWMZASR22, DBLP:journals/corr/abs-2402-05133}.
While these piecemeal approaches improve personalization on its own, there lacks a unified framework to bring them together. 
In this work, we bridge the gap by mirroring the dual-memory model in the human cognition process for more effective LLM personalization.

\paragraph{Datasets.}
The advancement of personalized LLMs has been hampered by a shortage of comprehensive benchmarks~\citep{tseng-etal-2024-two}. Existing ones predominantly target short‐context queries~\citep{DBLP:conf/cikm/LiZC20, salemi-etal-2024-lamp}, and some only contain 
user-level metadata~\citep{DBLP:journals/tiis/HarperK16, wu-etal-2020-mind}. 
These tasks, while useful, assess personalization in a rather shallow way, such as simple rating prediction for short movie reviews~\citep{ni-etal-2019-justifying} or capturing surface-level stylistic pattern in writing~\citep{salemi-etal-2024-lamp}. They overlook subtle dimensions of personalization, such as users' latent stance and evolving preferences during extended interactions. 
We also refer readers to~\Cref{sec:additional_literature_review} for additional discussions on the personalization evaluation.

\subsection{Memory Mechanism for LLM}

Decades of psychological research have converged on the following human memory components: sensory register, short-term memory, and long-term memory~\citep{AtkinsonShiffrin1968}. Regarding the durable long-term memory, further distinction has been made between \textit{episodic memory} and \textit{semantic memory}~\cite{tulving1972episodic,  tulving1985many}. Episodic memory refers to autobiographical events we can re-experience~\citep{annurev:/content/journals/10.1146/annurev.psych.53.100901.135114, CLAYTON2007R189}, e.g., recalling a specific conversation that happened last night. Semantic memory, on the other hand, refers to general facts and knowledge we have accumulated~\citep{saumier2002semantic, 10.1093/oxfordhb/9780195376746.013.0014}, such as knowing that NLP stands for Natural Language Processing. 
In this work, we posit that \textbf{the dual structure---episodic vs. semantic memories---is especially pertinent to LLM personalization}, as it mirrors the difference between remembering what happened in a particular interaction (\textit{episodes}), and knowing what is true about the users' opinions, beliefs, and preferences (\textit{semantics}).

Integrating memory into LLM-based systems quickly becomes a research frontier, as it holds the key to extending LLMs beyond fixed context windows, especially critical for LLM agents~\citep{zhang-etal-2024-llm-based, agent_memory_survey}. A standard implementation of episodic memory is retrieval-based: past interactions~\citep{DBLP:conf/uist/ParkOCMLB23} and external facts~\citep{DBLP:conf/iclr/YaoZYDSN023} are indexed in a database and fetched on demand. In contrast, semantic memory is mostly realized parametrically:\footnote{The use of memory in this work is to realize personalization, unlike existing ones which employ it for injecting factual \citep{fevry-etal-2020-entities} or moral \citep{zhang-etal-2024-moka} knowledge.} model’s parameters are updated by training on user data to embed user-level knowledge~\citep{DBLP:journals/corr/abs-2404-03565, DBLP:journals/corr/abs-2411-13405}.
Recent hybrid approaches attempt to combine these two by merely concatenating textual summaries with retrieved experiences~\citep{tan-etal-2024-democratizing, DBLP:conf/aaai/ZhongGGYW24, gupta-etal-2024-metareflection}, resulting in only superficial fusion. Recognizing the isolated usage and the shallow integration,
we formulate a more principled approach that enables deep information flow between episodic and semantic memories, which enables the successful use of the newly proposed {personalized thinking}. 

\section{CMV Dataset Construction}

Change My View (CMV) is a Reddit forum (\texttt{r/ChangeMyView}) where participants discuss to understand different viewpoints on various topics. CMV has been widely used for studies on argumentation~\citep{ji-etal-2018-incorporating, hua2019argument, DBLP:journals/corr/abs-2407-14829} and framing~\citep{DBLP:journals/corr/abs-2407-02637}. To our knowledge, we are the first to use CMV for LLM personalization, \textit{defining personalization as recommending the most persuasive reply for a given OP (original post)}. An evaluation example from our dataset is shown in~\Cref{fig:eval_item}.

\paragraph{\data Dataset Curation.} We obtained the raw CMV data (OPs, comments, and reply threads) from Academic Torrents.\footnote{\url{https://academictorrents.com/details/20520c420c6c846f555523babc8c059e9daa8fc5/}} We split the data chronologically: interactions from 2013--2022 form the \emph{historical engagement set} and those from 2023--2024 form the \emph{evaluation set}. The 2023 cutoff is chosen to better mitigate the data contamination issue---evaluation data have been part of the training corpus---since many open-weight models used in this study
have the knowledge cutoff in 2023~\citep{dong-etal-2024-generalization}. 
We restructure each interaction by flattening the original multi-branch structures into linear threads of (OP, direct reply, follow-ups). 
We discard any thread containing deleted contents or authors, marked with ``[deleted]'' or ``[removed]'', since they offer no helpful personalization signal. 

To convert conversations into a recommendation task, we exploit CMV’s \emph{delta} mechanism\footnote{OP authors award a ``$\Delta$'' to replies that change their view.} to label replies: A direct reply that receives a delta becomes a \emph{positive} example; all other direct replies under the same OP form the \emph{negative} pool. For the sake of simplicity, we only consider single-turn conversations and truncate all follow-ups. 

\paragraph{User Selection and Query Construction}
We restrict to \emph{active users} who awarded at least 10 deltas in the historical engagement set (2013--2022) and granted at least one delta in 2023--2024. This yields 56 authors. Each evaluation query contains an OP and one of its delta-awarded replies with non-delta replies to the same OP as negatives. Our initial evaluation set
comprises 327 queries from 56 OP authors.
We further filter data based on their difficulty level, with details in~\Cref{sec:appx_data_filter} to mitigate popularity heuristics~\citep{DBLP:conf/sigir/JiSZL20}.

\paragraph{Statistics.} Our final evaluation set includes $133$ queries by $41$ OP authors, supported by $7,514$ historical conversations published from 2013 to 2022. For the evaluation set (2023-2024), OP posts average $409$ tokens; positive and negative replies average $200.2$ and $105.8$ tokens, respectively. Each positive reply is paired with $47.5$ negatives on average ($6,317$ negatives total). In the historical engagement set, active authors have on average $28.1$ positive and $155.1$ negative conversations each.

\section{Memory Instantiation}
\label{sec:prelim}

Inspired by cognitive theories of memory~\citep{tulving1972episodic}, we investigate how different instantiations of episodic and semantic memories affect the LLM personalization. More specifically, we are interested in instantiating the memory-writing mechanism, i.e., how experiences are \textit{encoded} into memory, and the memory-reading mechanism, i.e., how that information is \textit{utilized} at test time. This study aims to provide insights into the \textbf{strengths and limitations} of various memory configurations.

\paragraph{Personalization with Dual-Memory.}
We adopt the dual-memory architecture, comprising episodic memory (EM) and semantic memory (SM), to define a \textbf{personalized LLM}, denoted as $\tilde{\mathcal{M}}$. The model processes an input query $x$ from user $a$ as follows:

\vspace{-8mm}
\begin{align}
\tilde{\mathcal{M}}(x)
= & \mathcal{M}(x;\textrm{EM}_a(x); \textrm{SM}_a(x)) \label{eq:overall} \\ 
= & \mathcal{M}(x ; \phi(x, \mathcal{H}(a)) ; \theta\oplus\Delta_{\mathcal{H}(a)})  \label{eq:decomposed}
\end{align}

\noindent $\mathcal{M}$ represents the base LLM with parameters $\theta$, $\mathcal{H}(a)$ denotes the historical engagements of user $a$, $\phi$ is the recall function for episodic memory, and $\Delta_{\mathcal{H}(a)}$ signifies the user-specific preference encoded in the personalized semantic memory. The operator $\oplus$ indicates the fusion of a base LLM and personalized adjustments.

For this set of preliminary experiments, we utilize \textsc{Llama-3.1}-8B~\citep{DBLP:journals/corr/abs-2407-21783} and \textsc{Qwen2.5}-7B~\citep{DBLP:journals/corr/abs-2412-15115}, for their representativeness. We conduct experiments on \data data, and see~\Cref{fig:eval_item} for a sample query.

\paragraph{Episodic Memory Instantiation.} 
The writing mechanism typically involves storing raw interaction data for efficiency and completeness. We thus focus on the reading mechanism, exploring several recall strategies, $\phi(\cdot)$: 
1) recall \textit{complete history}~\citep[i.a.,][]{DBLP:conf/nips/ShinnCGNY23}, 2) recall \textit{most recent histories}~\citep[i.a.,][]{wang2024user}, and 3) recall \textit{relevant histories}~\citep[i.a.,][]{DBLP:conf/uist/ParkOCMLB23}. Since full-history recall is intractable for long-context conversations, we focus our experiments on both recent and relevant recall. We also explore augmenting episodic memory with semantic memory–derived profile summaries ~\citep{DBLP:journals/corr/abs-2310-20081}, referred to as \textit{textual-summary augmentation (TSA).}\footnote{TSA concatenates the textual summary with recalled histories. Despite the hybrid memory usage, we classify it under \textit{Episodic Memory} per our adopted memory dichotomy.}

\begin{table}[t]
\centering
\resizebox{0.9\linewidth}{!}{%
\begin{tabular}{lrrr} \toprule
             & \multicolumn{1}{l}{Non-P} & Recent        & Relevant      \\ \midrule
Llama-3.1-8B & 26.58                    & 26.88 (26.67) & 25.68 (25.96) \\
Qwen2.5-7B   & 27.89                    & 25.51 (25.51) & 25.66 (26.18) \\ \bottomrule
\end{tabular}
}
\caption{
Aggregated results on \textbf{episodic memory} instantiation  (10 runs). Complete results refer to~\Cref{tbl:results_prelim_complte}. Parenthesized numbers represent textual-summary augmentation (TSA), which is beneficial for some cases. 
}
\vspace{-2mm}
\label{tbl:results_prelim_episodic}
\end{table}
\begin{table*}[t]
\centering
\resizebox{0.8\linewidth}{!}{%
\begin{tabular}{llrrrrrrrr} \toprule
             & \multirow{2}{*}{Non-P}     & \multicolumn{2}{c}{Input Only}         & \multicolumn{2}{c}{Fine Tuning}          & \multicolumn{2}{c}{Preference Tuning}    & \multicolumn{2}{c}{Textual}                                      \\  \cmidrule(lr){3-4} \cmidrule(lr){5-6} \cmidrule(lr){7-8} \cmidrule(lr){9-10}
             &                           & \multicolumn{1}{r}{NTP} & \multicolumn{1}{r}{CIG} & \multicolumn{1}{r}{O-FT} & \multicolumn{1}{r}{T-FT} & \multicolumn{1}{r}{DPO} & \multicolumn{1}{r}{SIMPO} & \multicolumn{1}{r}{HSumm} & \multicolumn{1}{l}{PKR (ours)} \\ \midrule
Llama-3.1-8B & \multicolumn{1}{r}{26.58} & 29.22                   & 29.79                   & 25.47                    & 31.24                    & 26.33                   & 24.45                     & 27.07                           & 26.62                          \\
Qwen2.5-7B   & \multicolumn{1}{r}{27.89} & 28.11                   & 28.41                   & 28.01                    & 30.20                    & 28.04                   & 17.37                     & 26.83                           & 27.02            \\ \bottomrule             
\end{tabular}
}
\vspace{-2mm}
\caption{
Average results of Hit@1, Hit@3, MRR, and DCG@3 on semantic memory configuration (10 runs)  Complete results refer to~\Cref{tbl:results_prelim_complte} where we additionally analyze the \textbf{time efficiency}. Non-P is a non-personalized baseline. Overall, the best configuration is to instantiate \textit{parametric semantic memory with task-oriented fine-tuning}, if the task information is available. Parametric semantic memory generally outperforms its textual counterpart, whereas the preference‐tuning approach delivers suboptimal results and thus deserves further investigation.}
\vspace{-2mm}
\label{tbl:results_prelim_semantic}
\end{table*}

\paragraph{Semantic Memory Instantiation.}
We first explore different instantiations of the memory-writing function, specifically focusing on deriving $\Delta_{\mathcal{H}(a)}$ by \textit{internalizing} information from user history $\mathcal{H}(a)$, i.e., encoding abstract concepts (e.g., preferences) into semantic memory. There are two forms of personalized semantic memory: parametric and textual forms. We provide a brief summary and \Cref{tbl:input_output} presents the input–output mappings for each instantiation.

\textbf{Parametric form}, encodes user preferences into the model's parameters. We examine several training objectives:

\begin{itemize}[leftmargin=1em,itemsep=0em,topsep=1pt,parsep=1pt,partopsep=1pt]
\item \textbf{Input-Only Training}: Suitable when human-written personalized outputs are unavailable~\citep{tan-etal-2024-democratizing}. Objectives include \textit{next token prediction (NTP)} and \textit{conditional input generation (CIG)}, e.g., generate a post based on the title.

\item \textbf{Fine-Tuning (FT)}: The most common practice to personalize model parameters~\citep{DBLP:journals/corr/abs-2404-03565, DBLP:journals/corr/abs-2411-13405, tan-etal-2024-democratizing}, and we have two variants: \textit{output-oriented FT (O-FT)} and \textit{task-oriented FT (T-FT)}, depending on whether end task information is handy.

\item \textbf{Preference Tuning}:  Alternative to RLHF, employs methods like \textit{DPO}~\citep{rafailov2023direct} and \textit{SIMPO}~\citep{meng2024simpo}, an efficient variant without the need for the reference model, to align model outputs with user preferences. Although RLHF has been used to learn user preferences~\citep{DBLP:journals/corr/abs-2402-05133}, its simpler alternative, preference tuning, remains largely unexplored for LLM personalization. 
\end{itemize}

\textbf{Textual form} represents user preferences as text, usually in the summary form. We explore:

\begin{itemize}[leftmargin=1em,itemsep=0em,topsep=1pt,parsep=1pt,partopsep=1pt]
\item \textit{Hierarchical Summarization (HSumm)}: hierarchically aggregates current interactions into concise summaries~\citep{DBLP:conf/aaai/ZhongGGYW24}.
\item \textit{Parametric Knowledge Reification (PKR)}: a novel method that leverages a model, trained on a user’s engagement history\footnote{In our experiments, we use \textit{T-FT} trained LoRAs.} to infer a concise profile summary. \textit{PKR} offers a speed gain over \textit{HSumm}.
\end{itemize}

During the \textbf{memory reading }process, as shown in~\Cref{eq:decomposed}, if semantic memory is in parametric form, the model parameters are adjusted as $\theta + \Delta_{\mathcal{H}(a)}$, i.e., $\oplus$ is matrix addition; if in the textual form, $\oplus$ is implemented as prefixing the generated profile summary to the input query $q$, i.e.,  $\tilde{\mathcal{M}}(x) =  \mathcal{M}([\Delta_{\mathcal{H}(a)}; x] ; \textrm{EM}_a(x); \theta)$ 

For instantiations that involve training, we utilize LoRA~\citep{hu2022lora} for its  efficiency and interpretability, allowing $\Delta_{\mathcal{H}(a)}$ as an abstract state to represent user-specific preferences and beliefs.

\paragraph{Discussion and Analysis.} Tables~\ref{tbl:results_prelim_episodic} and~\ref{tbl:results_prelim_semantic} present comprehensive results for episodic and semantic memory instantiations. \Cref{tbl:results_prelim_complte} also provides insights into the efficiency aspect. 

Our experiments reveal that episodic memory grounded in simple recency often outperforms a semantic‐similarity retrieval strategy---both in accuracy and speed---because the most recent interactions tend to be the strongest predictors of immediate user behavior. In contrast, semantic memory allows us to infer user preferences and latent traits even without task‐specific labels, as validated by the  improved performances achieved through \textit{input-only training}. The best performance is reached by the \textit{task fine-tuning (T-FT)}, which directly learns the mapping from the input query to the final desired outcome. Surprisingly, preference‐tuning methods underperform here, which deserves more investigation in the future.  Overall, using semantic memory (SM) alone generally leads to higher performance compared to using episodic memory (EM) alone. This suggests that semantic abstraction of user preferences and history might be more effective for personalization than simply recalling specific interactions.

It is important to emphasize that most of these memory instantiations have been examined individually in prior work, but never evaluated together on a common benchmark. To our knowledge, this study delivers the first comprehensive, head-to-head assessment of their personalization performance on long-context queries under a unified evaluation framework.

\section{Personalization with \framework}
\label{sec:method}

\Cref{sec:prelim} offers insights into the instantiation of episodic memory and semantic memory separately, which is a common practice in the literature. Only a few works attempt to combine the two, and those mostly operate in the textual space only~\citep{DBLP:journals/corr/abs-2310-20081,DBLP:conf/aaai/ZhongGGYW24}. To this end, we introduce \textsc{DUAL} (\Cref{sec:dual_framework}) to unify both memory types, so that \textit{the model can leverage detailed event histories alongside generalized user profiles}. Eventually, we introduce \framework (~\Cref{fig:framework}), by augmenting \textsc{DUAL} with \textit{personalized thinking} (\cref{sec:personalized_think}), which jointly leverages these memories to generate more faithful, user-aligned responses and exhibit richer personalized reasoning traces for improved interpretability.

\subsection{A Unified Framework with Cognitive Memory}
\label{sec:dual_framework}

This \textsc{DUAL} framework draws inspiration from the well-established cognitive theory of the dual-memory model~\citep{tulving1972episodic}.
To maintain efficiency during training and inference, we implement episodic memory via \textit{recency-based recall} and semantic memory via \textit{task-oriented fine-tuning (T-FT)}.  Importantly, our dual-memory framework is virtually compatible with all valid instantiation approaches, as confirmed in~\cref{sec:prelim}. Once instantiated, we freeze both memories.
 
At test time (right part of~\Cref{fig:framework}), we process each input query $x$ from an arbitrary user $a$ following~\Cref{eq:decomposed}. That is, we activate the corresponding LoRA matrices trained for the user $a$ to enable personalized semantic memory through parameters merging, $\theta + \Delta_{\mathcal{H}(a)}$. Next, we retrieve the most recent experiences for user $a$ from the episodic memory to form a context-aware input query, $x \oplus \phi(x, \mathcal{H}(a))$, where $\oplus$ denotes text concatenation. 

\subsection{Personalized Thinking}
\label{sec:personalized_think}
Slow thinking, demonstrated by long CoT methods like DeepSeek-R1~\citep{guo2025deepseek} and s1~\citep{muennighoff2025s1}, has shown promise, but its use in personalization is still in its infant stage. We are thus motivated to apply the slow thinking strategy to unlock personalized thinking

However, due to the fast thinking
training paradigm (i.e., direct mapping from input to output), we find that fine-tuned LLMs have been turned into a specialist model and overfitted to the target space, i.e., losing the generalist capability of generating meaningful intermediate thoughts when prompted. A common error is repetition of tokens.
To this end, we decide to unlock personalized thinking capabilities through training on \textit{synthesized personalized thoughts}.

\paragraph{Personalized Thoughts Generation} Capitalizing on the recent success of self-distillation~\citep{zhang2019your, pham2022revisiting, wang-etal-2023-self-instruct}, we design the following algorithm to produce intermediate thoughts and feed them back to the model itself for learning the personalized thinking process. We start by an LLM with instantiated \textit{parametric semantic memory}, i.e., $\tilde{\mathcal{M}}_{\textrm{SE}_\alpha}(\cdot) =  \mathcal{M}(\cdot;\emptyset; \textrm{SM}_a(\cdot))$

\begin{itemize}[leftmargin=1em,itemsep=0em,topsep=1pt,parsep=1pt,partopsep=1pt]
    \item Step 1 (Profile Generation): We prompt $\tilde{\mathcal{M}}_{\textrm{SE}_\alpha}(x)$ to generate a summary for a user, derived from the training on that user's history, following the same \textit{Parametric Knowledge Reification} approach as described in~\Cref{sec:prelim}.\footnote{Despite the model fails to generate meaningful thoughts, we find it still capable to generate meaningful summaries.}
    \item Step 2 (Review History Engagement): We convert each historical engagement into a query as in the evaluation format (\Cref{fig:eval_item}), and we prompt the $\tilde{\mathcal{M}}_{\textrm{SE}_\alpha}(x)$ to revisit all past engagements, and produce answers on them.
    \item Step 3 (Fast-thinking Filtering): We compare the produced answer with the ground truth answer, and then apply rejection sampling~\citep{zhu-etal-2023-solving, DBLP:journals/corr/abs-2308-01825} to keep the queries that the model is able to get right.
    \item Step 4 (Proxy LLM Initialization \& Reasoning): We follow the \textit{textual semantic memory} reading process in~\cref{sec:prelim} and use the summary generated by $\tilde{\mathcal{M}}_{\textrm{SE}_\alpha}(\cdot)$ in Step 1 to instantiate a proxy model, $\tilde{\mathcal{M}'}_{\textrm{SE}_\alpha}(\cdot) = \mathcal{M}(\cdot;\textrm{EM}_a(\cdot);\emptyset)$. Next, we conduct reverse engineering by providing the input query $x$ along with its answer to $\tilde{\mathcal{M}'}_{\textrm{SE}_\alpha}(\cdot)$. Using a prompt similar to~\Cref{fig:prime_prompt}, we instruct $\tilde{\mathcal{M}'}_{\textrm{SE}_\alpha}(\cdot)$ to generate structured intermediate thoughts in the same format as~\Cref{appx:fig:personalized_reasoning}, while requiring it to explicitly \textit{reiterate} the profile summary in its response.
    \item Step 5 (Slow-thinking Filtering): After obtaining the intermediate thoughts and predicted answer produced by the proxy LLM, we perform another round of rejection sampling to keep the ones where the final answer matches the ground truth.
\end{itemize}

\paragraph{Training:} After obtaining the synthesized personalized thoughts, we fine-tune $\tilde{\mathcal{M}}_{\textrm{SE}_\alpha}(\cdot)$ for a single epoch using cross-entropy loss. The input is still a plain query $q$, but the model is expected to generate both the personalized thinking trace (\textit{including the user's profile summary}\footnote{In our preliminary study, we find it generally advantageous to include profile summary in response.}) and the final answer.

Our work also draws a clear distinction from concurrent work~\citep{tsai-etal-2024-leveraging, tang2025think, zhang2025slow} on eliciting slow-thinking for LLM personalization. 
Unlike theirs, which focus solely on recommendation, we address diverse personalization tasks. Methodology-wise, \citet{tsai-etal-2024-leveraging} rely on a larger teacher model for reasoning path distillation, whereas we employ cost-effective self-distillation.  \citet{tang2025think, zhang2025slow} use virtual tokens,  i.e., a sequence of vectors, as intermediate steps while we are producing real tokens for the intermediate reasoning step, offering users interpretable reasoning traces.
% Third, the study in \citet{zhang2025slow} is limited to small-scale models (50M parameters) with its efficacy for larger models remain unverified, while we experiment with a diverse array of LLMs ranging from 3B to 14B.

\begin{table*}[t]
\centering
\vspace{-4mm}
\resizebox{0.85\linewidth}{!}{%
\begin{tabular}{lrrrrrrrrrr} \toprule
             & \multicolumn{2}{c}{Non-P}                           & \multicolumn{2}{c}{EM}                              & \multicolumn{2}{c}{SM}                              & \multicolumn{2}{c}{DUAL}                            & \multicolumn{2}{c}{PRIME}                           \\ \cmidrule(lr){2-3} \cmidrule(lr){4-5}  \cmidrule(lr){6-7} \cmidrule(lr){8-9}  \cmidrule(lr){10-11}
             & \multicolumn{1}{r}{Hit@3} & \multicolumn{1}{r}{Avg} & \multicolumn{1}{r}{Hit@3} & \multicolumn{1}{r}{Avg} & \multicolumn{1}{r}{Hit@3} & \multicolumn{1}{r}{Avg} & \multicolumn{1}{r}{Hit@3} & \multicolumn{1}{r}{Avg} & \multicolumn{1}{r}{Hit@3} & \multicolumn{1}{r}{Avg} \\ \midrule
Llama-3.2-3B & 38.65 & 26.44 & 37.89 & 26.27 & \textbf{43.61} & \textbf{30.25} & 41.95 & 28.87 & 42.93          & 29.95          \\
Llama-3.1-8B & 36.77 & 26.58 & 37.22 & 26.88 & 43.01          & 31.24          & 44.59 & 32.24 & \textbf{45.79} & \textbf{34.13} \\
Ministral-8B & 36.77 & 25.60 & 38.27 & 26.67 & \textbf{40.83} & 27.97          & \textbf{40.83} & 28.39 & 40.75          & \textbf{28.99} \\
Qwen2.5-7B   & 39.10 & 27.89 & 37.47 & 25.51 & 43.38          & 30.20          & 41.58 & 28.71 & \textbf{45.19} & \textbf{32.29} \\
Qwen2.5-14B  & 41.28 & 30.24 & 43.01 & 30.65 & 51.35          & 37.22          & 52.03 & 37.68 & \textbf{52.03} & \textbf{38.15} \\
Phi-4        & 41.50 & 29.63 & 42.93 & 30.31 & 42.63          & 31.09          & 43.98 & 32.61 & \textbf{47.29} & \textbf{35.15}   \\ \bottomrule   
\end{tabular}
}
\vspace{-2mm}
\caption{
Results on \data evaluation set (average of 10 runs). Avg is the aggregated metric of Hit@1, Hit@3, DCG@3, and MRR. Refer to~\Cref{tbl:CMV_complete_results} for detailed breakdown results, as well as performances of \textit{generic reasoners} (i.e., R1-distilled models) and retrieval-based \textsc{EM}. Best results for \textit{each} base model are in \textbf{bold}. Non-P is a non-personalized baseline. \framework performs better across the board, and is compatible with various base models. Note that we empirically do not include \textsc{EM} in \framework; For results on the \textsc{EM}-included \framework, refer to~\Cref{tbl:CMV_complete_results}.}
\label{tbl:results_CMV}
\end{table*}

\section{Experiments}

\paragraph{Datasets and Tasks.} We conduct a holistic evaluation of LLM personalization across four task types (ranking, classification, regression, and generation), and on both short- and long-context queries. To this end, we benchmark models on our curated \data to specifically probe long-context understanding. We also include a public LLM personalization benchmark, LaMP~\citep{salemi-etal-2024-lamp}, offering a testbed for all aforementioned tasks except for the ranking task. Specifically, we include LaMP-1 (citation identification), LaMP-2 (movie tagging), LaMP-3 (Amazon product rating),  LaMP-4 (news headline generation) and LaMP-5 (scholarly title generation), and remove LaMP-6 and 7.\footnote{We exclude LaMP-6 because it relies on a private dataset to which we have no access, and LaMP-7 because its GPT-3.5–generated labels do not truly represent real user behavior.} Dataset statistics are included in~\Cref{tbl:data_stats}, and \Cref{fig:eval_item} shows an evaluation example from \data dataset.  Evaluation metrics for each task are in~\Cref{sec:eval_metrics},

\paragraph{Setup.} We include recent, strong LLMs, showing promising results on various benchmarks. Specifically, we cover a diverse array of LLMs, ranging from mini LLMs (3B) to medium LLMs (14B), as shown in~\Cref{tbl:results_CMV} and discussed in~\Cref{appx:models}.

On LaMP benchmark, for fair comparison with the SOTA approach~\citep{tan-etal-2024-democratizing}, which is built upon \textsc{Llama2}-7B~\citep{DBLP:journals/corr/abs-2307-09288}, we only report performances based on \textsc{Llama-3.1}-8B.

\paragraph{Baselines and \framework Variants.} On both benchmarks, we compare  \framework with the non-personalized baseline, and \textit{generic reasoners} like \textsc{R1-Distill-Llama}. We also compare our approach with the SOTA system on LaMP tasks, OPPU~\citep{tan-etal-2024-democratizing}, which uses $100$× more data by training on vast non-target users' history before fine-tuning on a target user’s history.

Meanwhile, we compare \framework with its variants: episodic memory only (\textsc{EM}), semantic memory only (\textsc{SM}), and \textsc{DUAL}. For \framework variants, their memories are instantiated exactly as \framework.

\paragraph{Implementation Details.} \textsc{EM} uses 3 history interactions for CMV and 4 for LaMP benchmark. LoRAs used in \textsc{SM} are trained for up to 10 epochs for \textit{NTP} objective, 500 steps for preference tuning, and 3 epochs for other objectives (but 2 epochs for LaMP). For preference tuning, we pair one positive example with up to 10 negative examples under the same OP as the positive. 
At inference-time, \framework is allowed to generate up to $2,048$ tokens.
In our final implementation of \framework, we drop EM as it degrades performance (\Cref{tbl:CMV_complete_results}).

\begin{table}[t]
\hspace{-2mm}
\resizebox{1.05\linewidth}{!}{%
\begin{tabular}{lrrrrrrr} \toprule
Task  (Metric)                   & \multicolumn{1}{l}{Non-P} & \multicolumn{1}{l}{R1} & \multicolumn{1}{l}{\textsc{EM}} & \multicolumn{1}{l}{\textsc{SM}} & \multicolumn{1}{l}{\textsc{DUAL}} & \multicolumn{1}{l}{\framework} &  \multicolumn{1}{l}{SOTA} \\ \midrule
 LaMP1 (Acc) $\uparrow$    & 44.7                                 & 47.2                         & 49.6                              & 46.3                              & 52.8                                    & \textbf{54.5}             & 79.7              \\
 LaMP1 (F1) $\uparrow$    & 30.9                                 & 46.9                         & 45.7                              & 31.7                              & 52.9                                    & \textbf{54.5}             & 79.4            \\  \midrule
 LaMP2 (Acc) $\uparrow$   & 33.6                                 & 29.6                         & 43.6                              & 53.3                              & 50.0                                    & \textbf{54.3}             & 64.8             \\
 LaMP2 (F1) $\uparrow$    & 28.2                                 & 26.5                         & 34.4                              & 40.5                              & 39.1                                    & \textbf{42.7}             & 54.0             \\  \midrule
 LaMP3 (MAE) $\downarrow$   &.313                                &.366                        &.268                             &.214                             & \textbf{.188}                          &.223                     &.143            \\
 LaMP3 (RMSE) $\downarrow$   &.605                                &.620                        &.567                             &.482                             & \textbf{.453}                          &.491                     &.378            \\  \midrule
 LaMP4 (R-1) $\uparrow$   & 12.4                                 & 11.5                         & 13.8                              & 16.9                              & 18.6                                    & \textbf{18.8}             & 19.4               \\
 LaMP4 (R-L) $\uparrow$    & 11.0                                 & 10.1                         & 12.4                              & 15.2                              & 16.7                                    & \textbf{16.8}             & 17.5               \\ \midrule
LaMP5 (R-1) $\uparrow$   & 44.8                                 & 40.7                         & 47.0                              & 50.1                              & \textbf{52.2}                           & 47.3                      & 52.5              \\
LaMP5 (R-L) $\uparrow$   & 34.7                                 & 32.7                         & 38.5                              & 44.8                              &  \textbf{47.3}                    & 40.6                      & 47.3    \\     \bottomrule   
\end{tabular}
}
\vspace{-2mm}
\caption{\
Results on LaMP benchmark. Non-P is the non-personalized baseline, R1 denotes \textsc{R1-Distill-Llama}.  SOTA results, OPPU~\citep{tan-etal-2024-democratizing}, use $100$× more data of non-target users for training. Best performance among non-OPPUs is in \textbf{bold}. In general, personalized thinking in \framework leads to better results while the \textsc{DUAL} variant offers a competitive baseline.}
\vspace{-4mm}
\label{tbl:results_LaMP}
\end{table}

\section{Results and Analyses}

\subsection{Main results}
\label{sec:main_results}
Major results are included in ~\Cref{tbl:results_CMV}, and the full results (across all 5 metrics) can be found in~\Cref{tbl:CMV_complete_results}. Below are our major findings.

1) \textit{Generic Reasoning has limitations}: Enabling generic chain‐of‐thought often underperforms the non‐thinking baseline (see \Cref{tbl:CMV_complete_results}). The uncustomized reasoning trace merely scratches the surface, seeking broad answers rather than to-the-point, user‐specific responses. A detailed case study appears in \Cref{appx:case_study}.

2) \textit{Semantic Memory (\textsc{SM}) Beats Episodic Memory (\textsc{EM})}: Consistent with our major finding in~\Cref{sec:prelim}, \textsc{SM} alone generally outperforms \textsc{EM} alone, regardless of the model size or family.

3) \textit{\textsc{DUAL} Often Underperforms \textsc{SM} Alone}: Surprisingly, integrating both memory types without personalized thinking (\textsc{DUAL}) occasionally yields lower or comparable results than \textsc{SM} along. This suggests that potential conflicts between episodic and semantic memories could backfire if not properly mediated.

4) \textit{Model-agnostic Effectiveness}: \framework consistently enhances performance across all base models at different scales, illustrating that our \framework framework is robust and model-agnostic.

5) \textit{Personalized Thinking is Crucial}:
By augmenting \textsc{DUAL} with personalized thinking, \framework achieves superior performance over nearly all variants. This highlights the pivotal role of customized reasoning in improving personalization. 
Further, \Cref{appx:case_study} presents a case study comparing personalized and generic thinking traces, showcasing that our personalized thinking trace effectively adapts to the target user's profile.

\paragraph{Results on LaMP.}

LaMP is a public benchmark of short‐context queries that mainly tests surface‐level personalization (e.g., imitating writing style). Although \framework is designed to capture latent, evolving preferences, we are also interested in \framework's ability of handling short, simple queries. 

As shown in \Cref{tbl:results_LaMP}, the trends mirror those on \data: \textsc{SM} outperforms \textsc{EM}, and \text{DUAL} sometimes trails \textsc{SM} due to potential memory conflicts. Crucially, personalized thinking\footnote{Thinking prompt is shown in~\Cref{fig:prime_prompt_lamp}. \Cref{appx:fig:personalized_reasoning_lamp1} and \Cref{appx:fig:personalized_reasoning_lamp2} present personalized thinking traces on select LaMP tasks.} in \framework helps yield better results while \textsc{DUAL} is a competitive baseline for surface-level personalization tasks. This is inline with recent findings that overthinking might harm simple tasks~\citep{DBLP:journals/corr/abs-2503-16419}.
While PRIME surpasses all non-SOTA baselines, it remains behind the OPPU~\citep{tan-etal-2024-democratizing}, which is trained on 100× more data and includes other users’ histories. This cross-user training could violate privacy constraints in reality by exposing private data~\citep{kim2025personalized}. Given \framework's use of each user’s own history only, we deem the remaining gap acceptable.

\subsection{Further Analyses}

\paragraph{Train-free Personalized Thinking.} Cold‐start---performing personalized tasks with minimal history (e.g., $\leq$5 engagements)---remains challenging~\citep{DBLP:journals/corr/abs-2501-01945}. We thus decide to approach this challenge with our proposed \textit{personalized thinking} but under the training-free setting. Specifically, we prompt the \textsc{EM} variant with the thinking prompt (\Cref{fig:prime_prompt}). We also contrast it with vanilla \textsc{EM} using the standard prompt.

As shown from Figure~\ref{fig:analysis_1avg} to~\ref{fig:analysis_1mrr}, personalized thinking boosts all metrics except Hit@1 for all base models, compared to non-thinking counterparts. Despite trailing the trained \framework, it always outperforms other baselines including the strong generic reasoner (i.e., R1-distill LLMs), highlighting personalized thinking’s promise even without training.

\paragraph{Profile Replacement.} To evaluate the extent to which \framework faithfully leverages and ingest a user’s unique history, we perform a controlled ``profile‐replacement'' experiment: for each test query, we substitute the \textit{target user}’s engagement history---—both episodic (e.g., textual profile summary) and semantic memories (LoRA weights)---with that of  (i) the most similar user, (ii) a similar user, (iii) a mid-range user, or (iv) a dissimilar user.\footnote{Implementation details of substitutions see~\Cref{sec:appx_profile_replacement}.} We also include a \textit{Self} condition, retaining the user’s own history as a baseline.

\begin{figure}[t]
    \centering
    \includegraphics[width=0.42\textwidth]{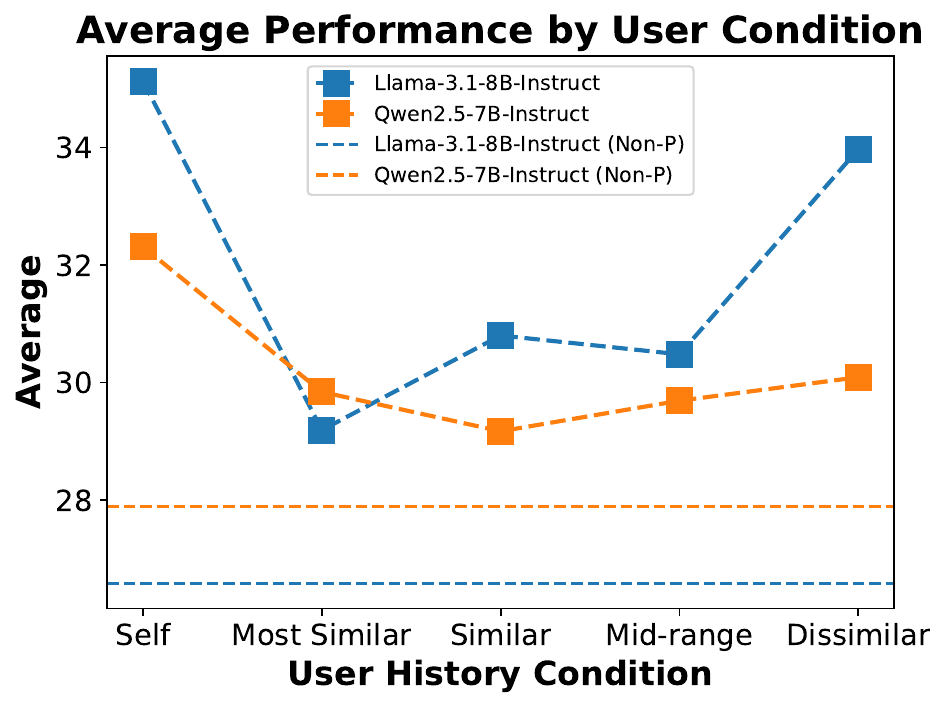}
    \vspace{-4mm}
    \caption{Average performance under five user-profile replacement conditions. Performance drops sharply when a target user's profile is replaced, confirming the faithfulness of \framework to user history. Non-P refers to non-personalized baseline.}
    \label{fig:analysis_1_change_author}
    \vspace{-6mm}
\end{figure}

We report average performance in~\Cref{fig:analysis_1_change_author} and detailed breakdown (e.g., Hit@k) in~\Cref{fig:analysis_1_change_author_appx}. 
For both evaluated models, performance is at peak when using the target user’s own profile (\textit{Self}). Replacement with any other user’s profile reduces performances, and more interestingly, the drop is steepest when the replaced profiles are similar to the original, and then partially recovers as the replacement profile becomes more divergent.

\textit{Why worst performance with similar profiles?} Our \framework learns fine-grained, user-specific preferences---effectively a dedicated bias towards which replies persuade that user most when evaluated on the \data dataset. Noticeably, due to the data scarcity of limited active users on the CMV forum, two users with superficially similar posting histories may differ sharply in which replies they find compelling. Replacement with similar profiles thus misleads the model way more than any other incorrect profiles: \framework confidently applies the \textit{wrong preferences}.

\textit{Why performance rebounds?}
When a replaced profile is quite different from the original, the loaded personalization signal, e.g., LoRA weights, is weak or irrelevant to the input query. \framework then falls back on its general, pre-trained commonsense knowledge for interence, without exploiting the strong but unaligned personalized cues, injected in the learning stage~(\Cref{sec:personalized_think}). As such, the model regains accuracy. Moreover, the improved performance over the non-personalized baseline also showcases the value of thinking mechanism for personalization tasks.

Overall, our observation confirms that \framework's reasoning and predictions depend critically on \textit{correct user history}, and cannot be explained by simple bandwagon or popularity heuristics~\citep{DBLP:conf/sigir/JiSZL20}. This further highlights that \framework faithfully captures dynamic, user‐specific preferences.

\section{Conclusion}
Inspired by the cognitive dual‐memory model, we first systematically study different memory instantiations and then propose \framework, a unified framework that integrates episodic and semantic memory mechanisms. We further augment \framework with a novel {personalized thinking} capability, yielding more accurate, user‐aligned responses and richer reasoning traces. To assess long‐context personalization, we introduce the \data dataset and conduct extensive experiments, which demonstrate the effectiveness of both \framework and personalized thinking. Finally, our further analysis confirms that \framework shows strong fidelity to each user’s unique history.

\section*{Acknowledgments}
This work was supported in part by the Air Force Office of Scientific Research
under grant FA9550-22-1-0099, and computational resources and services provided by Advanced Research Computing (ARC), a division of Information and Technology Services (ITS) at the University of Michigan, Ann Arbor. 
We thank the reviewers of ARR for their useful comments. 

\section*{Limitations}

\paragraph{Evaluation benchmarks.} In this work, we have included two evaluation benchmarks, aiming to cover a diverse array of tasks, genres and domains. Yet, these two benchmarks cannot comprehensively represent the entire spectrum. For example, recent research efforts venture into long-form personalization~\citep{DBLP:journals/corr/abs-2407-11016}. For future research, we plan to extend \framework to more applications, and examine its true generalizability in the wild.  

\paragraph{Model Scales.} In this study, we evaluate a diverse set of models ranging from 3$B$ to 14$B$ parameters. Owing to budget constraints, we do not extend our experiments to larger-scale models.
Nonetheless, we expect our findings to hold on such models, as suggested by extrapolation from the trends observed in~\cref{tbl:results_CMV}. We leave the direct application of of \framework to larger models in the future work.

\paragraph{GPU resources.} The base LLMs used in this work are of 3 to 14 billions parameters. 
It is thus more time-consuming than traditionally small models like BERT \citep{devlin-etal-2019-bert} at inference time, which in turn results in a higher carbon footprint.
Specifically, we run each base LM on 1 single NVIDIA A40 or NVIDIA L40 with significant CPU and memory resources. The combined inference time for each LLM on the three benchmarks ranges from 10 to 20 hours, depending on the configurations.

\setcounter{table}{0}
\setcounter{figure}{0}
\renewcommand{\thefigure}{A\arabic{figure}}
\renewcommand{\thetable}{A\arabic{table}}

\clearpage

\appendix

% \section{Example Appendix}
% \label{sec:appendix}

\section{Implementation Details}
\subsection{Models Used for Experiments on \data}
\label{appx:models}

For all base models, we use their instruction-fine-tuned versions for experiments. We have provided model cards in the footnotes.

\begin{itemize}[leftmargin=2em,itemsep=0em,topsep=1pt,parsep=1pt,partopsep=1pt]
    \item \textbf{Mini LLM}: \textsc{Llama-3.2}-3B~\cite{meta2024llama3_2}\footnote{meta-llama/Llama-3.2-3B-Instruct}
    \item \textbf{Small LLM}: \textsc{Llama-3.1}-8B~\citep{DBLP:journals/corr/abs-2407-21783},\footnote{meta-llama/Llama-3.1-8B-Instruct} \textsc{Qwen2.5}-7B~\citep{DBLP:journals/corr/abs-2412-15115},\footnote{Qwen/Qwen2.5-7B-Instruct} \textsc{Ministral}-8B~\citep{mistral2024ministraux}\footnote{mistralai/Ministral-8B-Instruct-2410}.
    \item \textbf{Medium LLM}: \textsc{Qwen2.5}-14B~\citep{DBLP:journals/corr/abs-2412-15115},\footnote{Qwen/Qwen2.5-7B-Instruct } \textsc{phi-4}~\citep{DBLP:journals/corr/abs-2412-08905}\footnote{microsoft/phi-4}
\end{itemize}

\subsection{Evaluation Metrics} 
\label{sec:eval_metrics}
For our \data benchmark, considering it is a ranking task~\citep{manning2008introduction}, we adopt Hit@1, Hit@3, DCG@3 and MRR. For LaMP, we follow the official metrics~\citep{salemi-etal-2024-lamp},
and use accuracy and F1-score for classification tasks (LaMP-1 and LaMP-2), MAE and RMSE for the ordinal regression task (LaMP-3), and ROUGE-1 and ROUGE-L~\citep{lin-2004-rouge} for text generation tasks (LaMP-4 and LaMP-5). Note that all metrics are the higher the better, except for RMSE and MAE used for the LaMP-3.

\section{Additional Literature Review on Evaluation Challenge}
\label{sec:additional_literature_review}
Although benchmarks like PRISM~\citep{DBLP:journals/corr/abs-2404-16019} and Empathetic Conversation~\citep{DBLP:journals/corr/abs-2205-12698} offer a testbed for long-context query evaluation, their evaluation relies on generic metrics, e.g., ROUGE~\citep{lin-2004-rouge}, or uses LLM‐as‐a-judge~\citep{DBLP:conf/nips/ZhengC00WZL0LXZ23}. The former only measures surface-level similarity, while the latter demands extensive prompt engineering~\citep{dong-etal-2024-llm, DBLP:conf/iui/SzymanskiZEL0M25} and still falls short of truly \textit{imitating} individual users' preferences~\citep{DBLP:conf/nips/JiangXZHZ023}, as the models do not consistently hold the target user's persona.
To address these gaps, we introduce a new CMV‐based benchmark focusing on long‐form, persuasion‐driven recommendation tasks, enabling direct and objective assessment of LLM personalization without relying on proxy judgments.

\section{\data Data Filtering by Difficulty}
\label{sec:appx_data_filter}
To ensure queries demand personalization rather than commonsense reasoning or popularity heuristics~\citep{DBLP:conf/sigir/JiSZL20}, we apply two instruction-tuned small yet powerful LLMs, i.e., \textsc{Llama-3.1}-8B~\citep{DBLP:journals/corr/abs-2407-21783} and  \textsc{Qwen2.5}-7B~\citep{DBLP:journals/corr/abs-2412-15115}, on each query\footnote{We only consider 9 sampled negatives to form a query, the same setting as in~\cref{sec:prelim}.} without providing user history. We perform 10 runs per model, computing Hit@1 and Hit@3. We retain queries with Hit@1 $\leq0.3$ and Hit@3 $\leq0.5$, removing 15 authors and 194 queries to focus on challenging personalization items.

\begin{table}[t]
\centering
\resizebox{0.9\linewidth}{!}{%
\begin{tabular}{lrrrr} \toprule
\multicolumn{1}{l}{Tasks}  & \#Q & |Q| & \#History & Output  Format            \\ \midrule
\data    & 133                     & 1561.4                  & 183.2                         & ranking             \\ \hdashline[5pt/4pt]
LaMP-1 & 123                     & 29.0                    & 317.5                         & 2-way class         \\ 
LaMP-2 & 3,302                   & 48.6                    & 55.6                          & 15-way class        \\
LaMP-3 & 112                     & 183.9                   & 959.8                         & {[}1, 2, 3, 4, 5{]} \\
LaMP-4 & 6,275                   & 18.2                    & 270.1                         & short generation    \\
LaMP-5 & 107                     & 161.9                   & 442.9                         & short generation   \\ \bottomrule
\end{tabular}
}
\caption{
Basic statistics of evaluation sets. \#Q indicates the number of queries. |Q| is token-based input query length, excluding template tokens. \#History tells the number of historical engagements per user.}
\label{tbl:data_stats}
\end{table}

\begin{figure*}[t]
    \centering
    \includegraphics[width=0.75\textwidth]{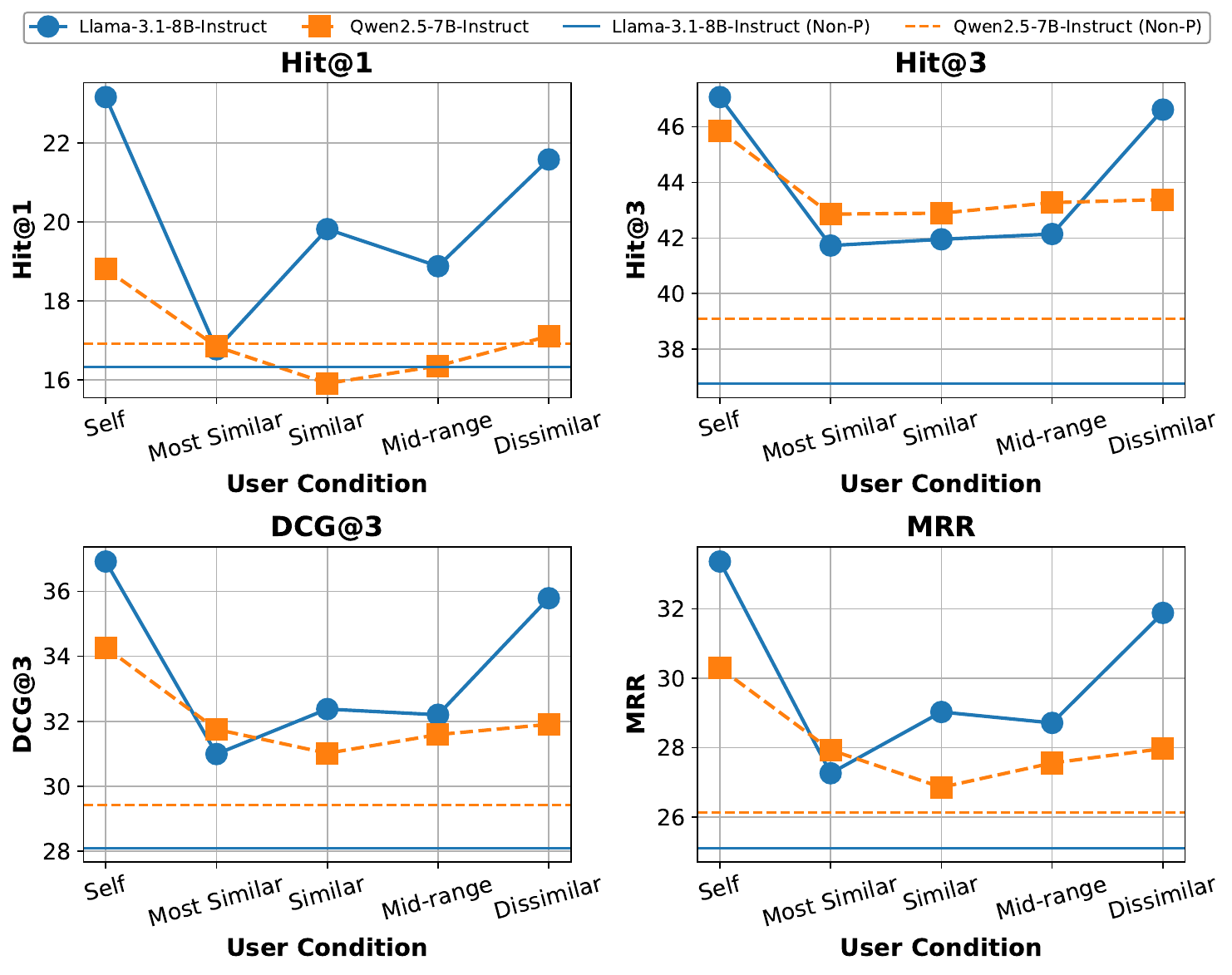}
    \caption{
    Hit@1, Hit@3, DCG@3 and MRR performances under five user-profile replacement conditions--self, most similar, similar, mid-range and dissimilar. Importantly, all metrics drop when the profile is replaced, confirming the sensitivity of \framework to user history, showing that \framework indeed captures the dynamic personalization rather than just bandwagon biases. Non-P refers to non-personalized baseline.}
    \label{fig:analysis_1_change_author_appx}
    \vspace{-4mm}
\end{figure*}

\begin{table*}[t]
\centering
\resizebox{0.9\linewidth}{!}{%
\begin{tabular}{llrrrrrrr} \toprule
Model        & Instantiation                       & \multicolumn{1}{l}{Hit@1} & \multicolumn{1}{l}{Hit@3} & \multicolumn{1}{l}{DCG@3} & \multicolumn{1}{l}{MRR} & \multicolumn{1}{l}{Avg} & W. Efficiency    &  R. Efficiency  \\ \midrule
\multicolumn{9}{c}{No Personalization}                                                                                                                                               \\ \hdashline[5pt/4pt]
Llama-3.1-8B & \multirow{2}{*}{Baseline}     & 16.32                     & 36.77                     & 28.10                      & 25.11                   & 26.58               & \multirow{2}{*}{N/A}                 & \multirow{2}{*}{N/A}    \\
Qwen2.5-7B   &                               & 16.91                     & 39.10                      & 29.43                     & 26.13                   & 27.89                   \\ \midrule
\multicolumn{9}{c}{Episodic Memory (EM)}                                                                                                                                             \\ \hdashline[5pt/4pt]
Llama-3.1-8B & \multirow{2}{*}{Recent}       & \textbf{16.62}            & 37.22                     & \textbf{28.36}            & \textbf{25.33}          & \textbf{26.88}          & \multirow{2}{*}{Fastest} & \multirow{2}{*}{Slow} \\
Qwen2.5-7B   &                               & 13.91                     & 37.47                     & 27.10                      & 23.57                   & 25.51                   \\
Llama-3.1-8B & \multirow{2}{*}{Relevant}     & 16.17                     & 35.41                     & 27.00                        & 24.12                   & 25.68        & \multirow{2}{*}{Fastest} & \multirow{2}{*}{Slower}            \\
Qwen2.5-7B   &                               & 13.23                     & 38.50                      & 27.36                     & 23.56                   & 25.66                   \\ \hdashline[5pt/4pt]
Llama-3.1-8B & \multirow{2}{*}{Recent+PKR}   & \textbf{16.62}            & 36.84                     & 28.10                      & 25.10                    & 26.67       &    \multirow{2}{*}{Medium}              & \multirow{2}{*}{Slower}        \\
Qwen2.5-7B   &                               & 14.29                     & 37.07                     & 27.05                     & 23.62                   & 25.51                   \\
Llama-3.1-8B & \multirow{2}{*}{Relevant+PKR} & 15.64                     & 36.32                     & 27.45                     & 24.41                   & 25.96    & \multirow{2}{*}{Medium}              & \multirow{2}{*}{Slowest}              \\
Qwen2.5-7B   &                               & 13.76                     & \textbf{39.02}            & 27.88                     & 24.07                   & 26.18                   \\ \midrule
\multicolumn{9}{c}{Semantic Memory (SM)}                                                                                                                                             \\ \hdashline[5pt/4pt]
Llama-3.1-8B & \multirow{2}{*}{NTP}          & 17.44                     & 41.20                      & 30.93                     & 27.31                   & 29.22         & \multirow{2}{*}{Fast}                & \multirow{2}{*}{Fast}            \\
Qwen2.5-7B   &                               & 16.84                     & 39.55                     & 29.71                     & 26.34                   & 28.11                   \\
Llama-3.1-8B & \multirow{2}{*}{CIG}          & 17.74                     & 41.95                     & 31.56                     & 27.92                   & 29.79        & \multirow{2}{*}{Fast}                & \multirow{2}{*}{Fast}            \\
Qwen2.5-7B   &                               & 16.77                     & 40.23                     & 30.05                     & 26.57                   & 28.41                   \\ \hdashline[5pt/4pt]
Llama-3.1-8B & \multirow{2}{*}{Output FT}    & 14.66                     & 36.47                     & 26.99                     & 23.75                   & 25.47     & \multirow{2}{*}{Medium-Fast}         & \multirow{2}{*}{Fast}                 \\
Qwen2.5-7B   &                               & 16.54                     & 39.85                     & 29.58                     & 26.08                   & 28.01                   \\
Llama-3.1-8B & \multirow{2}{*}{Task FT}      & \textbf{19.62}            & 43.01                     & \textbf{32.96}                     & \textbf{29.36}                   & \textbf{31.24}         & \multirow{2}{*}{Medium}              & \multirow{2}{*}{Fast}          \\ 
Qwen2.5-7B   &                               & 16.99                      & \textbf{43.38}            & 32.15            & 28.28          & 30.20         \\ \hdashline[5pt/4pt]
Llama-3.1-8B & \multirow{2}{*}{DPO}          & 15.41                     & 37.37                     & 27.89                     & 24.64                   & 26.33  &  \multirow{2}{*}{Slowest}             & \multirow{2}{*}{Fast}               \\
Qwen2.5-7B   &                               & 16.77                     & 39.55                     & 29.61                     & 26.22                   & 28.04                   \\
Llama-3.1-8B & \multirow{2}{*}{SIMPO}        & 14.21                     & 34.81                     & 25.89                     & 22.88                   & 24.45       & \multirow{2}{*}{Slow}                & \multirow{2}{*}{Fast}              \\
Qwen2.5-7B   &                               & 10.08                     & 24.66                     & 18.44                     & 16.30                    & 17.37                   \\ \hdashline[5pt/4pt]
Llama-3.1-8B & \multirow{2}{*}{HSumm}  & 16.32                     & 37.89                     & 28.62                     & 25.44                   & 27.07      & \multirow{2}{*}{Slowest}             & \multirow{2}{*}{Medium}               \\
Qwen2.5-7B   &                               & 15.04                     & 38.80                      & 28.50                      & 24.97                   & 26.83                   \\
Llama-3.1-8B & \multirow{2}{*}{PKR}          & 16.69                     & 36.39                     & 28.12                     & 25.26                   & 26.62        & \multirow{2}{*}{Medium}              & \multirow{2}{*}{Medium}           \\
Qwen2.5-7B   &                               & 15.34                     & 39.02                     & 28.63                     & 25.08                   & 27.02               \\ \bottomrule  
\end{tabular}
}
\caption{Complete results of the preliminary study where we study the strengths and limitations of various memory configurations. Results are the average of 10 runs. Recent/Relevant+PKR are effectively \textit{hybrid approaches} using both episodic and textual semantic memories, but we decide to place them under the Episodic Memory (EM) category per out adopted \textit{memory dichotomy}. \textit{W. Efficiency} refers to memory writing or memory instantiation efficiency. For episodic memories, the writing time is the index creation time cost, which is extremely fast, compared to semantic memory writing. For semantic memory, we determine the efficiency label based on the train flops. For example, given a history of 15 engagements, the train flops of NSP/CIG is around 1e+16, while that of DPO is almost 1e+17. R. Efficiency measures the time overhead of both memory reading and the subsequent inference step. This overhead grows linearly with the number of retrieved past interactions, and increases further if a textual profile summary is prepended. In contrast, parametric semantic memories incur minimal inference cost, since all it needs to process is the input query without worrying about the past interaction retrieval.}
\label{tbl:results_prelim_complte}
\vspace{-6mm}
\end{table*}

\begin{table*}[t]
\centering
\small
\vspace{-6mm}
\resizebox{0.7\linewidth}{!}{%
\begin{tabular}{lrrrrr} \toprule
Model & \multicolumn{1}{c}{Hit@1} & \multicolumn{1}{c}{Hit@3} & \multicolumn{1}{c}{DCG@3} & \multicolumn{1}{c}{MRR} & \multicolumn{1}{c}{Avg} \\ \midrule
\multicolumn{6}{c}{No Personalization}                                                                                                     \\ \hdashline[5pt/4pt]
Llama-3.2-3B          & 14.51                     & 38.65                     & 28.09                     & 24.49                   & 26.44                   \\
Llama-3.1-8B          & 16.32                     & 36.77                     & 28.10                     & 25.11                   & 26.58                   \\
Ministral-8B          & 14.36                     & 36.77                     & 27.27                     & 24.00                   & 25.60                   \\
Qwen2.5-7B            & 16.91                     & 39.10                     & 29.43                     & 26.13                   & 27.89                   \\
Qwen2.5-14B           & 19.40                     & 41.28                     & 31.77                     & 28.51                   & 30.24                   \\
Phi-4                 & 17.97                     & 41.50                     & 31.27                     & 27.77                   & 29.63                   \\ \hdashline[5pt/4pt]
DeepSeek-R1-Llama-3.1-8B & 13.61                     & 36.77                     & 26.68                     & 23.64                   & 25.18                   \\
DeepSeek-R1-Qwen2.5-7B   & 13.08                     & 33.76                     & 24.70                     & 21.91                   & 23.36                   \\
DeepSeek-R1-Qwen2.5-14B  & 17.97                     & 44.66                     & 32.96                     & 28.99                   & 31.15                   \\ \midrule
\multicolumn{6}{c}{\textsc{EM} (Recency)}                                                                                                    \\ \hdashline[5pt/4pt]
Llama-3.2-3B          & 14.74                     & 37.89                     & 27.94                     & 24.50                   & 26.27                   \\
Llama-3.1-8B          & 16.62                     & 37.22                     & 28.36                     & 25.33                   & 26.88                   \\
Ministral-8B          & 15.11                     & 38.27                     & 28.36                     & 24.95                   & 26.67                   \\
Qwen2.5-7B            & 13.91                     & 37.47                     & 27.10                     & 23.57                   & 25.51                   \\
Qwen2.5-14B           & 18.42                     & 43.01                     & 32.40                     & 28.76                   & 30.65                   \\
Phi-4                 & 17.74                     & 42.93                     & 32.13                     & 28.42                   & 30.31                   \\ \midrule
\multicolumn{6}{c}{\textsc{EM} (Retrieval)}                                                                                                  \\ \hdashline[5pt/4pt]
Llama-3.2-3B          & 14.59                     & 37.97                     & 27.93                     & 24.49                   & 26.25                   \\
Llama-3.1-8B          & 16.17                     & 35.41                     & 27.00                     & 24.12                   & 25.68                   \\
Ministral-8B          & 14.96                     & 37.82                     & 28.05                     & 24.70                   & 26.38                   \\
Qwen2.5-7B            & 13.23                     & 38.50                     & 27.36                     & 23.56                   & 25.66                   \\
Qwen2.5-14B           & 18.87                     & 41.35                     & 31.74                     & 28.43                   & 30.10                   \\
Phi-4                 & 17.89                     & 43.01                     & 32.29                     & 28.61                   & 30.45                   \\ \midrule
\multicolumn{6}{c}{\textsc{SM}}                                                                                                            \\ \hdashline[5pt/4pt]
Llama-3.2-3B          & 17.22                     & 43.61                     & 32.25                     & 27.91                   & 30.25                   \\
Llama-3.1-8B          & 19.62                     & 43.01                     & 32.96                     & 29.36                   & 31.24                   \\
Ministral-8B          & 15.34                     & \textbf{40.83}            & 29.78                     & 25.94                   & 27.97                   \\
Qwen2.5-7B            & 16.99                     & 43.38                     & 32.15                     & 28.28                   & 30.20                   \\
Qwen2.5-14B           & 23.38                     & 51.35                     & 39.16                     & 34.99                   & 37.22                   \\
Phi-4                 & 19.85                     & 42.63                     & 32.65                     & 29.24                   & 31.09                   \\ \midrule
\multicolumn{6}{c}{\textsc{DUAL}}                                                                                                          \\ \hdashline[5pt/4pt]
Llama-3.2-3B          & 16.09                     & 41.95                     & 30.78                     & 26.65                   & 28.87                   \\
Llama-3.1-8B          & 20.15                     & 44.59                     & 34.06                     & 30.16                   & 32.24                   \\
Ministral-8B          & 16.24                     & 40.83                     & 30.08                     & 26.40                   & 28.39                   \\
Qwen2.5-7B            & 15.71                     & 41.58                     & 30.66                     & 26.90                   & 28.71                   \\
Qwen2.5-14B           & 23.76                     & 52.03                     & 39.59                     & 35.34                   & 37.68                   \\
Phi-4                 & 21.58                     & 43.98                     & 34.13                     & 30.76                   & 32.61                   \\ \midrule
\multicolumn{6}{c}{\textsc{EM}-included \framework}                                                                                                          \\ \hdashline[5pt/4pt]
Llama-3.2-3B          & 16.47                     & \textbf{44.44}            & \textbf{32.54}            & \textbf{28.38}          & \textbf{30.46}          \\
Llama-3.1-8B          & 22.33                     & 43.46                     & 34.29                     & 31.08                   & 32.79                   \\
Ministral-8B          & 16.47                     & 39.25                     & 29.47                     & 25.94                   & 27.78                   \\
Qwen2.5-7B            & 15.11                     & 41.13                     & 30.04                     & 26.18                   & 28.12                   \\
Qwen2.5-14B           & 24.14                     & 50.08                     & 39.00                     & 35.19                   & 37.10                   \\
Phi-4                 & 21.95                     & \textbf{48.12}            & \textbf{36.93}            & 33.08                   & 35.02                   \\ \midrule
\multicolumn{6}{c}{\framework}                                                                                                           \\ \hdashline[5pt/4pt]
Llama-3.2-3B          & \textbf{17.29}            & 42.93                     & 31.81                     & 27.78                   & 29.95                   \\
Llama-3.1-8B          & \textbf{22.56}            & \textbf{45.79}            & \textbf{35.87}            & \textbf{32.28}          & \textbf{34.13}          \\
Ministral-8B          & \textbf{17.14}            & 40.75                     & \textbf{30.81}            & \textbf{27.26}          & \textbf{28.99}          \\
Qwen2.5-7B            & \textbf{19.47}            & \textbf{45.19}            & \textbf{34.16}            & \textbf{30.35}          & \textbf{32.29}          \\
Qwen2.5-14B           & \textbf{24.29}            & \textbf{52.03}            & \textbf{40.17}            & \textbf{36.09}          & \textbf{38.15}          \\
Phi-4                 & \textbf{23.01}            & 47.29                     & \textbf{36.93}            & \textbf{33.37}          & \textbf{35.15}          \\ \bottomrule
\end{tabular}
}
\caption{Full results on \data evaluation set (average of 10 runs). Avg is the aggregated metric of Hit@1, Hit@3, DCG@3, and MRR. Generic reasoners' performances are reported based on DeepSeek-R1-distilled models. Best results for \textit{each} base model are in \textbf{bold}. \framework performs better across the board, and is compatible with various base models. For EM, we include both recency- and retrieval-based memory instantiation, to help provide a more comprehensive understanding. For our proposed \framework framework, we have two variants: \textsc{EM}-included \framework is the exact implementation of~\Cref{sec:method}, which integrates \textsc{DUAL} with personalized thinking capability; In the final version, simply dubbed \framework, the EM component is dropped because we empirically observe that \textsc{DUAL} often underperforms \textsc{SM} alone (\Cref{sec:main_results}).}
\vspace{10mm}
\label{tbl:CMV_complete_results}
\end{table*}

\begin{table*}[ht]
\centering
\small
\vspace{10mm}
\begin{tabular}{%
    >{\raggedright\arraybackslash}p{2cm}  
    >{\raggedright\arraybackslash}p{8cm}    
    >{\raggedright\arraybackslash}p{4cm}   
}
\toprule
               & Input                                                                                                                                                             & Output                         \\
\midrule
NTP            & "author": \{author\}. "title": \{title\}. "body": \{body\}<EOS>                                                                    & author": \{author\}. "title": \{title\}. "body": \{body\}   \\[6pt]
CIG            & For the topic "\{title\}", the author "\{author\}" states:                                                                         & \{body\}                       \\[6pt]
Output-FT (O-FT)           & The author, "\{author\}", has engaged with users on the Change-My-View subreddit across various original posts (OPs). Based on the author"s preference and engagement patterns, generate a persuasive response that is highly likely to change their viewpoint on the following post. "title": \{title\}. "body": \{body\}  & "reply": \{positive\_reply\}     \\[6pt]
Task-FT  (T-FT)         & The author, "\{author\}", has engaged with users on the Change-My-View subreddit across various original posts (OPs) and is seeking alternative opinions to alter their viewpoint.\newline\newline Currently, the author is creating a new OP titled "\{title\}", with the content:\newline***\{body\}.***\newline From the "candidate replies" JSON file below, select the best reply (using "option ID") that best challenges the author"s view.\newline\{candidates\}  & {[}"\{option ID\}"{]}               \\[6pt]
Preference Tuning    & "author": \{author\}. "title": \{title\}\newline"body": \{body\}                                                                   & "reply": \{positive\_reply\}  / "reply": \{negative\_reply\}   \\
\bottomrule
\end{tabular}
\caption{Input-output mappings for training different instantiations of parametric semantic memory. In the context of a single-turn CMV conversation, there are always the following fields: \textit{title}, \textit{body}, \textit{author} and \textit{reply}. If a reply receives a $\Delta$, then it is a positive reply; otherwise, it is a negative reply. Such pair can directly support the preference tuning. However, in reality, it is not always the case we have the access to aforementioned items. If the reply is unavailable, one may resort to \textbf{NTP} or \textbf{CIG}. For \textbf{NTP}, the output is essentially the left-shifted input. If output is available, one may utilize fine-tuning paradigm such as \textbf{O-FT} or \textbf{T-FT}, where the latter will be preferred if we are able to know the task information. For \textbf{T-FT}, we convert raw data into the desired task format following the prompt shown in~\Cref{fig:CMV_prompt}.
Note, \{$\cdot$\} is a placeholder which is to be replaced with concrete content.}
\label{tbl:input_output}
\vspace{15mm}
\end{table*}

\section{Implementation Details of Profile Replacement Study}
\label{sec:appx_profile_replacement}

\paragraph{Author Index.} We build each user’s profile by concatenating their OP titles and contents\footnote{Including users’ delta-awarding histories did not substantially change performance trends.}. We then encode each profile with a sentence transformer model, \textsc{all-MiniLM-L6-v2}, to obtain author-level embeddings, which form our author index.

\paragraph{Substitution.} At test time, for each query we replace the target user’s profile with that of a \textit{replacement user}, chosen according to one of four replacement conditions below.  That is, we swap in the replacement user’s textual profile summary (episodic memory) and load their LoRA weights (semantic memory), swap out the original target user's memories, and then carry out the evaluation as follows.

We consider four replacement conditions:
\begin{itemize}[leftmargin=1em,itemsep=0em,topsep=1pt,parsep=1pt,partopsep=1pt]
\item \textit{Most Similar}: Replace with the user whose posting history is most similar to the original.
\item \textit{Similar}: Average performance when replacing with the 3rd- and 5th-most similar users.
\item \textit{Mid-range}: Average performance when replacing with the 10th- and 20th-most similar users (20 is the median in our \data dataset of 41 active users where one of them is the original target user).
\item \textit{Dissimilar}: Average performance when replacing with the least similar and the 10th-least similar users.
\end{itemize}

\section{Case study of Thinking Trace}
\label{appx:case_study}
This section provides a brief overview on the thinking traces by \framework (\Cref{appx:fig:personalized_reasoning}), i.e., personalized thinking, and by \textsc{R1-Distill-Llama} (\Cref{appx:fig:generic_reasoning}), which is a generic reasoner. 

Specifically, our personalized thinking explicitly emphasizes on the \textit{author's historical sensitivity to methodological rigor}, grounding its selection of option F (``data used to support their conclusion is flaw'') in a nuanced understanding of the author's past engagements. This showcases how our thinking traces implements personalization.  In contrast, the generic reasoner (\textsc{R1-Distill-Llama}) superficially mentions methodological issues in option F but ultimately prioritizes other responses (A, C, and D) based on \textit{broader counter-evidence and direct contradictions}. Thus, \textsc{R1-Distill-Llama}'s reasoning is weakened by a lack of personalization, failing to adequately weigh methodological precision aligned with the author's long-term behavior and patterns.

\begin{figure*}[t]
    \centering
    \includegraphics[width=0.96\textwidth]{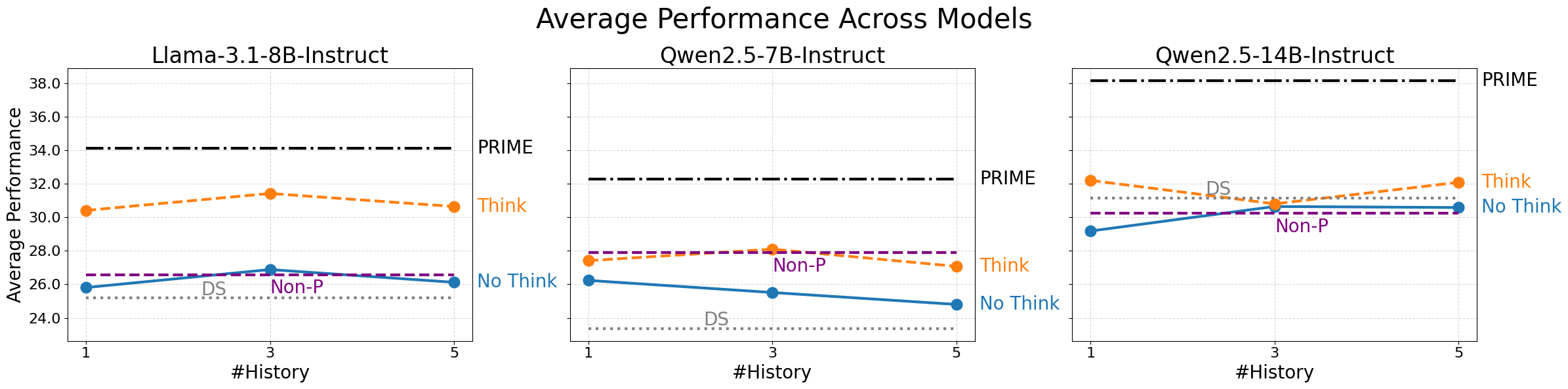}
    \caption{Average performance for Train-free Personalized Thinking study. \textit{Think} refers to our train-free thinking approach, i.e., prompt the \textsc{EM} variant with the thinking prompt (\Cref{fig:prime_prompt}). \textit{No Think} is the non-thinking baseline where we prompt vanilla \textsc{EM} with the standard prompt. \textit{Non-P} is the no personalization baseline. \textit{DS} denote the generic reasoner version of each base model. Concretely, we use \textsc{DeepSeek-R1-Distill-Llama-8B}, \textsc{DeepSeek-R1-Distill-Qwen-7B}, and \textsc{DeepSeek-R1-Distill-Qwen-14B}, respectively.
    }
    \label{fig:analysis_1avg}
\end{figure*}

\begin{figure*}[t]
    \centering
    \includegraphics[width=0.96\textwidth]{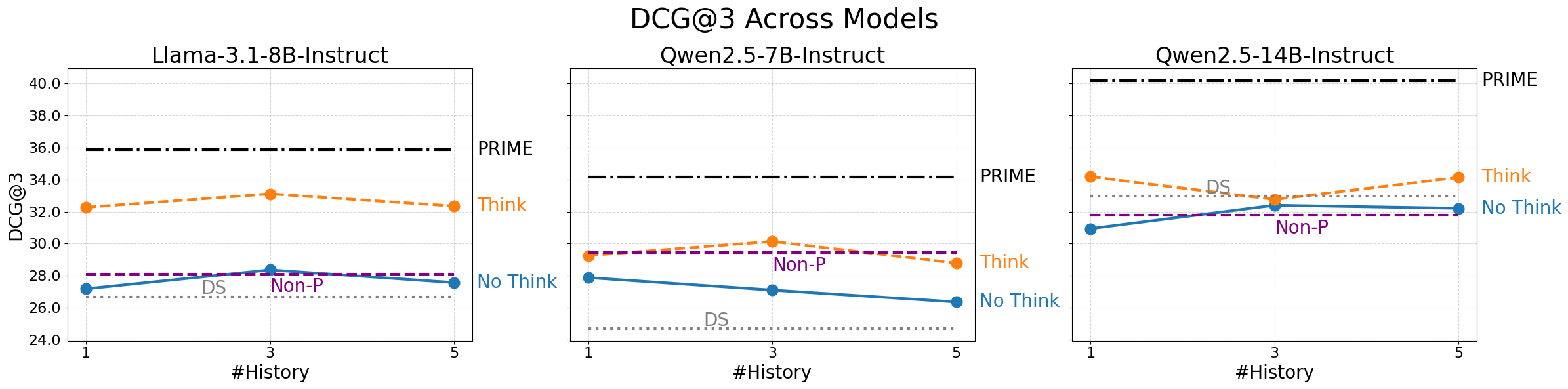}
    \caption{DCG@3 metric for Train-free Personalized Thinking study. Legend explanations are in~\Cref{fig:analysis_1avg}.
    }
    \label{fig:analysis_1dcg}
    \vspace{10mm}
\end{figure*}

\begin{figure*}[t]
    \centering
    \includegraphics[width=0.96\textwidth]{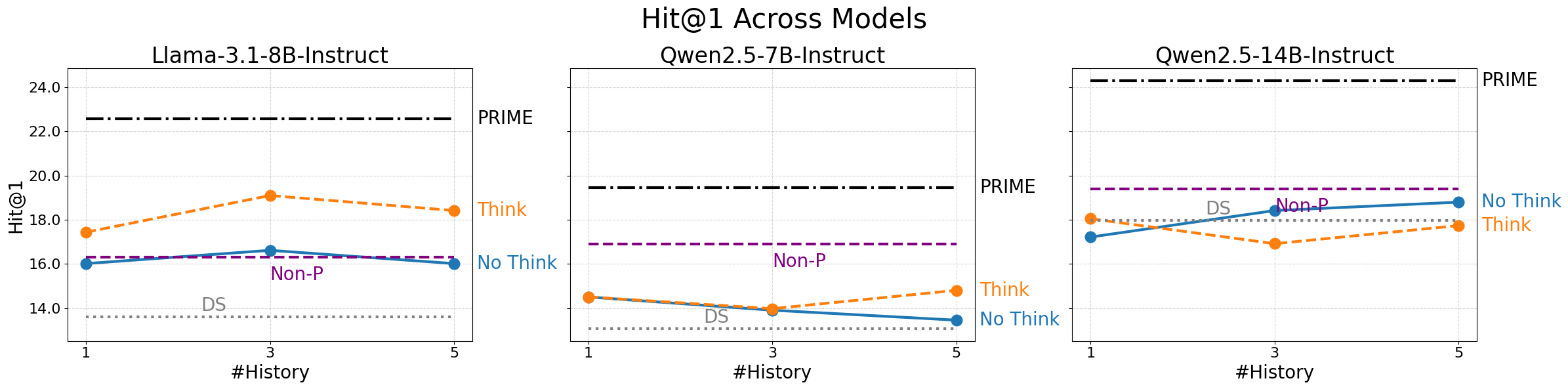}
    \caption{Hit@1 metric for Train-free Personalized Thinking study. Legend explanations are in~\Cref{fig:analysis_1avg}.
    }
    \label{fig:analysis_1h1}
\end{figure*}

\begin{figure*}[t]
    \centering
    \includegraphics[width=0.96\textwidth]{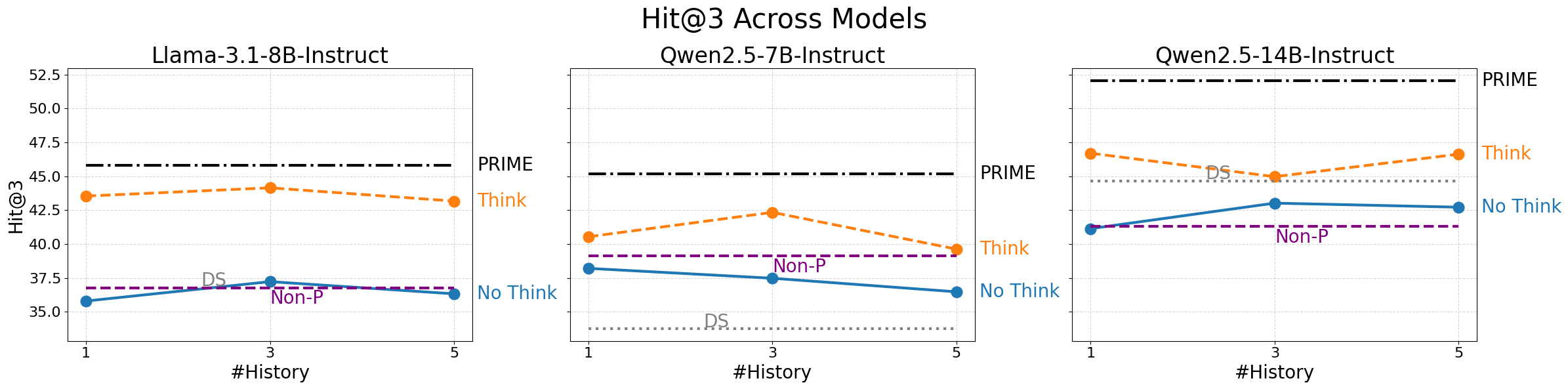}
    \caption{Hit@3 metric for Train-free Personalized Thinking study. Legend explanations are in~\Cref{fig:analysis_1avg}.
    }
    \label{fig:analysis_1h3}
\end{figure*}

\begin{figure*}[t]
    \centering
    \includegraphics[width=0.96\textwidth]{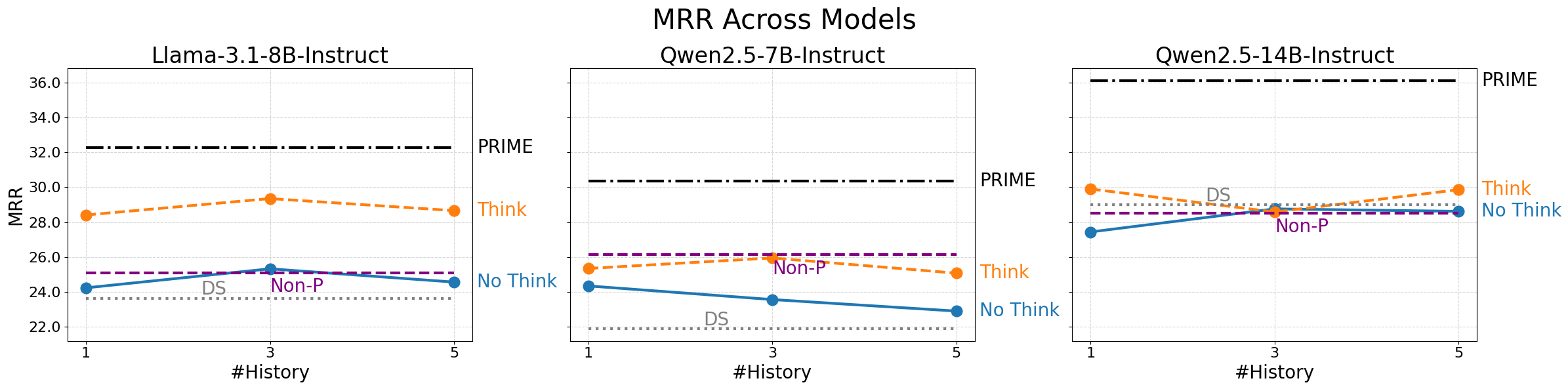}
    \caption{MRR metric for Train-free Personalized Thinking study. Legend explanations are in~\Cref{fig:analysis_1avg}.
    }
    \label{fig:analysis_1mrr}
\end{figure*}

\clearpage

\begin{figure*}[t]    
  \centering
  \small
\begin{tcolorbox}[
  colback=white,
  colframe=black,
  coltitle=white,
  colbacktitle=black,
  title=\small\bfseries Evaluation Template,
  boxrule=0.8pt,
  arc=1mm,
  left=1em,
  right=1em,
  top=1em,
  bottom=1em
]
The author, \texttt{\{AUTHOR\}}, has engaged with users on the Change-My-View subreddit across various original posts (OPs) and is seeking alternative opinions to alter their viewpoint.

Currently, the author is creating a new OP titled
\begin{lstlisting}
     {OP TITLE}
\end{lstlisting}
with the following content:

\begin{lstlisting}
     {OP CONTENT}
\end{lstlisting}

From the \texttt{candidate replies} JSON file below, select the top 3 replies (using \texttt{option ID}) that best challenge the author’s view. Rank them from most to least compelling.  

\medskip
\begin{lstlisting}
[ 
  { 'option ID': '...','challenger':'...','reply':'...'}, 
  { 'option ID': '...','challenger':'...','reply':'...'},
]
\end{lstlisting}

-----------------------------------------------------------------------------------------------------------------------------------------------

\medskip
Output a valid JSON array of \texttt{``option ID"} strings representing the selected replies. Each element must be a double-quoted string. The response should contain nothing but the JSON array and end with ``\#END''.
\end{tcolorbox}
\caption{Standard prompt template for \data evaluation, used by no-personalization baseline and generic reasoners. The template contains the \textit{input template} (top) and \textit{output instruction} (bottom). For \textsc{EM} variants that requires the access to recalled histories, the histories are prepended to the current OP, i.e., right before the word ``\textit{Currently}''.
This design is inspired by the recent work on using structured outputs~\citep{zhang-etal-2024-narrative}, and we also prompt LLMs to produce outputs in JSON format to ease the post-hoc processing.}
\label{fig:CMV_prompt}
\end{figure*}

\begin{figure*}[t]       
  \centering
  \small
  
\begin{tcolorbox}[
  colback=white,
  colframe=black,
  coltitle=white,
  colbacktitle=black,
  title=\small\bfseries Evaluation Query,
  boxrule=0.8pt,
  arc=1mm,
  left=1em,
  right=1em,
  top=1em,
  bottom=1em
]
The author, \texttt{kingpatzer}, has engaged with users on the Change-My-View subreddit across various original posts (OPs) and is seeking alternative opinions to alter their viewpoint.

Currently, the author is creating a new OP titled
\begin{displayquote}
\enquote{CMV: Those who attribute gun ownership rates as the cause of the problem of gun violence in terms of criminal gun deaths are not merely mistaken; they are disingenuous}
\end{displayquote}
with the following content:

\begin{displayquote}
The data has been clear for a very long time: the relationship between guns and gun homicides doesn't show any strong correlation.

I have personally taken the cause-of-death data from https://wonder.cdc.gov/, grouping results by year and state, and selecting \emph{Homicide, Firearm} as the cause of death. I then matched that data to the per-capita gun-ownership statistics by state from the ATF, as reported by Hunting Mark (https://huntingmark.com/gun-ownership-stats/).

A standard correlation analysis between firearm homicide rates per 100{,}000 and per-capita gun ownership yields an $r^2$ of 0.079 (no meaningful correlation). A similar global analysis by nation gives an $r^2$ of 0.02\ldots

The only way to associate gun ownership with gun violence is by including suicides by firearm, which I argue is disingenuous. We don’t count suicide by hanging as “rope violence” when discussing strangulation, nor overdoses as “drug violence,” etc.
\end{displayquote}

From the \texttt{candidate replies} JSON file below, select the top 3 replies (using \texttt{option ID}) that best challenge the author’s view. Rank them from most to least compelling.  

\medskip
\begin{lstlisting}
[ 
  { 'option ID': '...','challenger':'...','reply':'...'}, 
  { 'option ID': '...','challenger':'...','reply':'...'},
]
\end{lstlisting}
-----------------------------------------------------------------------------------------------------------------------------------------------

\medskip
Output a valid JSON array of \texttt{``option ID"} strings representing the selected replies. Each element must be a double-quoted string. The response should contain nothing but the JSON array and end with ``\#END''.
\end{tcolorbox}
\caption{A sample evaluation query from our \data dataset, formatted using the standard prompt (\Cref{fig:CMV_prompt}). Due to space limit, we are unable to show the entire  \texttt{candidate replies} list, but we have provided a candidate reply in~\Cref{fig:eval_item_gt}.}
\label{fig:eval_item}
\end{figure*}

\begin{figure*}[t]      
  \centering
\begin{tcolorbox}[
  colback=white,
  colframe=black,
  boxrule=0.8pt,
  arc=1mm,
  left=1em,
  right=1em,
  top=1em,
  bottom=1em,
  title=\small\bfseries Positive Candidate Reply (option F)
]
\small
\begin{itemize}[leftmargin=1em]
  \item \textbf{Option ID}: \texttt{option F}
  \item \textbf{Challenger}: \texttt{An-Okay-Alternative}
  \item \textbf{Reply}:
    \begin{quote}
      Your data for gun-ownership is weapons per capita, not individual gun owners per capita.  
      If we’re testing access to guns vs.\ gun homicides, someone who owns 30 guns wouldn’t be 30× as likely to commit a homicide.

      Secondly, the ATF list includes only specially regulated firearms (e.g.\ fully automatic weapons, short-barreled shotguns) and NFA items bought by law enforcement.

      Thus, the data omits common firearms, is biased by collectors and hobbyists, and is confounded by law-enforcement purchases.

      Overall, this calculation is weak evidence against a gun–homicide correlation and doesn’t imply critics are dishonest.
    \end{quote}
\end{itemize}
\end{tcolorbox}
\caption{Positive candidate reply to the evaluation query shown in~\Cref{fig:eval_item}, which receives $\Delta$ awarded by the OP author.}
\label{fig:eval_item_gt}
\end{figure*}

\vspace{-16mm}
\begin{figure*}[t]   
  \centering
{\small
\begin{tcolorbox}[
  colback=white,
  colframe=black,
  boxrule=0.8pt,
  arc=1mm,
  left=1em,
  right=1em,
  top=1em,
  bottom=1em,
  title=\small\bfseries Personalized Thinking Prompt (Output Instruction Part Only)
]
You are an AI assistant designed with \textbf{bionics-inspired episodic memory} capabilities. Your objective is to approach the task by emulating human-like episodic recall, drawing from past interactions, contextual understanding, and nuanced reasoning to deliver informed and thoughtful judgments.

\textbf{Treat the author's past conversations as episodic memories that guide your reasoning and decision-making throughout the process. Prioritize capturing the user's values and patterns from past interactions and integrating these into your reasoning.}

\section*{Instructions}
\begin{enumerate}
  \item \textbf{Ingest Author History:}
    \begin{itemize}
      \item Leveraging your trained semantic memory, extract and synthesize insights from the author's prior interactions. Summarize the author's past patterns, preferences, values, and beliefs to establish an episodic memory.
      \item Write a concise summary of history conversations within \texttt{<user experience>} tags. This summary should serve as your episodic memory for later steps.
    \end{itemize}
  \item \textbf{Summarize the New OP:}
    \begin{itemize}
      \item Review the content of the new OP carefully, identifying salient events, major arguments, core themes, and the author's explicit and implicit viewpoints.
      \item Write a concise summary of the new OP within \texttt{<OP summary>} tags.
    \end{itemize}
  \item \textbf{Sketch an Outline:}
    \begin{itemize}
      \item Combine insights from your episodic memory (\texttt{<user experience>}) with the context from the new OP (\texttt{<OP summary>}). Conduct reasoning that incorporates the author's past preferences and patterns to create a strategic outline for how to challenge or respond to the author's viewpoint.
      \item Highlight the most important points or questions for challenging the author's view and encapsulate these in a concise outline within \texttt{<sketch outline>} tags.
    \end{itemize}
  \item \textbf{Evaluate Candidate Replies:}
    \begin{itemize}
      \item Analyze each candidate reply from the provided JSON file in terms of strength, relevance, and weaknesses. Base your evaluations on your episodic memory, OP reasoning, and the outlined strategy.
      \item Present this evaluation as a dictionary within \texttt{<analysis>} tags, e.g., \texttt{\{'option A': [analysis], 'option B': [analysis], \dots\}}.
    \end{itemize}
  \item \textbf{Reflect and Rank Top Replies:}
    \begin{itemize}
      \item Reflect on and integrate all insights to determine the most compelling replies—those engaging the author's view and providing reasoned, respectful, and novel insights.
      \item Identify the top three replies by option ID, ranking them from highest to lowest compellingness. Include your concise reflection within \texttt{<reflection>} tags.
    \end{itemize}
  \item \textbf{Answer and Conclude:}
    \begin{itemize}
      \item Output your selection as a valid JSON array of strings within \texttt{<answer>} tags, e.g., \texttt{["option ID", "option ID", "option ID"]}.
      \item End your response immediately with \#END.
    \end{itemize}
\end{enumerate}

\textbf{Output Format:}
\begin{verbatim}
<user experience>[concise user experience summary]</user experience>
<OP summary>[concise OP summary]</OP summary>
<sketch outline>[concise sketched outline]</sketch outline>
<analysis>{'option A': [analysis], 'option B': [analysis], ...}</analysis>
<reflection>[concise reflection]</reflection>
<answer>["option ID","option ID","option ID"]</answer>
#END
\end{verbatim}
\end{tcolorbox}
}
\caption{The \textit{output instruction} part of our personalized thinking prompt, designed for \data evaluation. The complete prompt used by \framework is the concatenation of the \textit{input template} part as shown in~\Cref{fig:CMV_prompt} and the \textit{output instruction} part shown here.}
\label{fig:prime_prompt}
\end{figure*}

\clearpage

\vspace{-16mm}
\begin{figure*}[t]   
  \centering
{\small
\begin{tcolorbox}[
  colback=white,
  colframe=black,
  boxrule=0.8pt,
  arc=1mm,
  left=1em,
  right=1em,
  top=1em,
  bottom=1em,
  title=\small\bfseries Personalized Thinking Prompt for LaMP
]
The user, \texttt{\{AUTHOR\}}, is performing a \texttt{\{TASK\}} task to \texttt{\{TASK DESCRIPTION\}}.\\
Currently, 
\begin{lstlisting}
     {CONTENT}
\end{lstlisting}

-----------------------------------------------------------------------------------------------------------------------------------------------

You are an AI assistant designed with \textbf{bionics-inspired episodic memory} capabilities. Your objective is to approach the task by emulating human-like episodic recall, drawing from past interactions, contextual understanding, and nuanced reasoning to deliver informed and thoughtful judgments.

\textbf{Treat the author's past conversations as episodic memories that guide your reasoning and decision-making throughout the process. Prioritize capturing the user's values and patterns from past interactions and integrating these into your reasoning.}

\section*{Instructions}
\begin{enumerate}
  \item \textbf{Ingest User History:}
    \begin{itemize}
      \item Leveraging your trained semantic memory, extract and synthesize insights from the author's prior interactions. Summarize the author's past patterns, preferences, values, and beliefs to establish an episodic memory.
      \item Write a concise summary of history engagement within \texttt{<user experience>} tags. This summary should serve as your episodic memory for the later steps.
    \end{itemize}
  \item \textbf{Summarize the New Query:}
    \begin{itemize}
      \item Review the content of the new query carefully, identifying salient aspects and key objectives reflected in the user's input.
      \item Write a concise summary of the new query within \texttt{<query summary>} tags.
    \end{itemize}
  \item \textbf{Reflect:}
    \begin{itemize}
      \item Combine insights from your episodic memory (\texttt{<user experience>}), the context from the new query (\texttt{<query summary>}), and the task description (\texttt{<task description>}).
      \item \texttt{<task description> \texttt{\{TASK\}}: \texttt{\{DETAILED TASK DESCRIPTION\}} </task description>}
      \item Based on the task description, conduct reasoning that incorporates the user's past preferences and patterns to reflect on the new query.
      \item Include the concise reflection within \texttt{<reflection>} tags.
    \end{itemize}
  \item \textbf{Answer and Conclude:}
    \begin{itemize}
      \item \texttt{\{OUTPUT INSTRUCTION\}}
      \item Enclose this final answer within \texttt{<answer>} tags.
      \item End your response immediately with \#END.
    \end{itemize}
\end{enumerate}

\textbf{Output Format:}
\begin{verbatim}
<user experience>[concise user experience summary]</user experience>
<query summary>[concise query summary]</query summary>
<reflection>[concise reflection]</reflection>
<answer>[final answer]</answer>
#END
\end{verbatim}
\end{tcolorbox}
}
\caption{Personalized thinking prompt designed for LaMP evaluation. \{$\cdot$\} are placeholders to be replaced with corresponding contents (\Cref{tbl:lamp_prompt}).}
\label{fig:prime_prompt_lamp}
\end{figure*}

\begin{table*}[t]
\centering
\resizebox{1.0\linewidth}{!}{%
\begin{tabular}{lp{0.4\linewidth}p{0.6\linewidth}l}
\hline
Task                                 & Task Description                                   & Detailed Task Description                                                                                                                                       & Output Instruction                                                              \\
\hline
Citation Identification & Select citations for articles authored by the user & The citation behavior of researchers is dependent on their interests and is commonly used to evaluate and develop personalized systems for recommending papers. & Respond with only ``{[}1{]}'' or ``{[}2{]}'' and end with ``\#END''. \\
Movie Tagging           & Assign tags to movies based on the user's preferences & Users tagging behavior for media such as movies and books are known to be idiosyncratic and depends on their understanding of the tag and the aspects of the item they focus on. & Respond with only a valid tag and end with ``\#END''. \\
Product Rating          & Provide scores for purchased products based on the user's preferences & Product reviews commonly express a nuanced set of user preferences for a product and which in turn determine their rating for the product. & Respond with only a valid rating and end with ``\#END''. \\
News Headline Generation & Write headlines for news articles authored by the user & Authors writing displays distinct stylistic elements influenced by both personal and social factors. & Respond with only the generated headline and end with ``\#END''. \\
Scholarly Title Generation & Generate titles for scholarly articles authored by the user & Authors writing displays distinct stylistic elements influenced by both personal and social factors. & Respond with only the generated title and end with ``\#END''.\\
\hline
\end{tabular}
}
\vspace{-2mm}
\caption{Overview of five personalized tasks in LaMP benchmark~\citep{salemi-etal-2024-lamp}, their task descriptions, detailed task descriptions, and output instructions.}
\label{tbl:lamp_prompt}
\end{table*}

\clearpage

\begin{figure*}[t]    
  \centering
\begin{tcolorbox}[
  colback=white,
  colframe=black,
  boxrule=0.8pt,
  arc=1mm,
  left=1em,
  right=1em,
  top=1em,
  bottom=1em,
  title=\small\bfseries Personalized Thinking Trace
]
\small
\textbf{User Experience:} The author, \texttt{kingpatzer}, has engaged with users on the Change-My-View subreddit across various original posts (OPs), seeking alternative opinions to alter their viewpoint. In past discussions, they’ve shown openness to different perspectives and a willingness to consider data and evidence about gun ownership and gun violence. However, they can become defensive when challenged and sometimes dismiss opposing viewpoints.

\medskip
\textbf{OP Summary:} The author is creating a new OP titled
\begin{displayquote}
CMV: Those who attribute gun ownership rates as the cause of the problem of gun violence in terms of criminal gun deaths are not merely mistaken; they are disingenuous
\end{displayquote}
They argue that available data show no strong correlation between gun ownership rates and gun homicides, and that including firearm suicides is disingenuous—comparing it to “rope violence” or “drug violence” when discussing strangulation or overdoses.

\medskip
\textbf{Sketch Outline:}

To challenge the author's view, we need to address the following points:

1. The author's interpretation of the data: We need to examine the author's methodology and data analysis to see if it is sound and if it supports their conclusion.
2. The inclusion of suicide by guns: We need to discuss the relevance of including suicide by guns in the data set and why it is not disingenuous to do so.
3. The comparison to other causes of death: We need to compare the author's approach to other causes of death, such as drug overdoses, to see if it is consistent and fair.
4. The author's tone and language: We need to address the author's tone and language, which some may perceive as dismissive and condescending.

Key questions to ask:

* What is the author's definition of "disingenuous" and how does it apply to including suicide by guns in the data set?
* Is the author's methodology sound and does it support their conclusion?
* How does the author's approach to gun ownership and gun violence compare to other causes of death?
* Why is the author using a tone and language that some may perceive as dismissive and condescending?

\medskip
\textbf{Analysis (option F):}

This reply challenges the author's view by pointing out that the data used to support their conclusion is flawed, and that the author's methodology is not sound. The reply also highlights the importance of considering the social and cultural context in which gun ownership and gun violence occur.

\medskip
\textbf{Reflection:}

Based on the analysis, the top 3 replies that best challenge the author's view are:

1. Option F: This reply challenges the author's view by pointing out that the data used to support their conclusion is flawed, and that the author's methodology is not sound. The reply also highlights the importance of considering the social and cultural context in which gun ownership and gun violence occur.
2. Option D: This reply challenges the author's view by pointing out that there is a strong link between gun ownership and impulsive suicide, and that including suicide by guns in the data set is not disingenuous. The reply also highlights the importance of considering the cultural and social context in which gun ownership and gun violence occur.
3. Option C: This reply provides a detailed and well-researched argument that challenges the author's view, pointing out that there is a significant correlation between gun ownership and gun violence. The reply also highlights the importance of considering the social and cultural context in which gun ownership and gun violence occur.

These replies are the most compelling because they engage with the author's perspective, provide reasoned and respectful arguments, and challenge the author's view in a thoughtful and nuanced way.

\medskip
\textbf{Answer:}
\[
  \bigl[\text{"option F"},\ \text{"option D"},\ \text{"option C"}\bigr]
\]
\end{tcolorbox}
\caption{Personalized thinking trace generated by our \framework to the evaluation query shown in~\Cref{fig:eval_item}. Due to space limit, we only show the analysis for option F here (refer to~\ref{appx:fig:personalized_reasoning_complete_analysis} for the complete analysis), but we keep the presentation of all the other components intact.}
\label{appx:fig:personalized_reasoning}
\end{figure*}

\clearpage

\begin{figure*}[t]   
  \centering
\begin{tcolorbox}[
  colback=white,
  colframe=black,
  boxrule=0.8pt,
  arc=1mm,
  left=1em,
  right=1em,
  top=1em,
  bottom=1em,
  title=\small\bfseries Generic Reasoning Trace by \textsc{R1-Distill-Llama}
]
\small
\textbf{Think:}

Alright, I need to figure out which of the provided replies best challenge the author’s viewpoint. The author argues that there’s no strong correlation between gun ownership rates and gun homicides, and that including suicides in the data is disingenuous.

Looking at the replies, I should focus on those that either provide counterarguments, point out flaws in the author’s methodology, or present evidence contradicting their claims.

Option A: This reply points out that ignoring suicides is misleading because many suicides could be prevented by restricting gun access. It also mentions that while suicides by other means are controlled, there’s no such regulation for guns. This directly challenges the author’s exclusion of suicides, making it a strong point.

Option B: This uses an analogy about other causes of death to question the author’s approach. It’s a clever way to highlight that the author’s reasoning might be inconsistent, suggesting the argument is flawed. This is compelling because it questions the validity of the author’s methodology.

Option C: This reply cites a study that found a positive correlation between gun ownership and firearm homicides. It’s strong evidence against the author’s claim, making it a solid challenge.

Option D: This argues that there’s a correlation between gun ownership and gun deaths, including suicides, and criticizes the author’s exclusion of suicides. It’s a direct counterargument, supporting the idea that the author is being disingenuous.

Option E: This reply references a global study showing a link between gun ownership and gun violence, which directly contradicts the author’s data. It’s a good point because it uses external evidence to challenge the author.

Option F: This critique points out methodological issues with the author’s data, such as not measuring individual gun owners and including restricted firearms. It’s detailed and highlights the limitations of the author’s approach, making it a strong challenge.

Option G: This reply calls the author’s statement absurd, stating that gun homicides are clearly correlated with gun presence. It’s a strong, straightforward counterargument.

Option H: This argues that while the author excludes suicides, measures are taken for other causes, but not for gun suicides. It’s relevant because it highlights the inconsistency in excluding suicides, supporting the idea of disingenuousness.

Option I: This defends the author’s character, saying disagreement doesn’t make someone disingenuous. It’s more of a defense rather than a direct challenge, so it’s less compelling.

Option J: This points out that the global study the author cited was flawed and incorrect. It’s a good challenge because it questions the validity of the author’s global data.

After evaluating, the top three replies that most effectively challenge the author’s viewpoint are A, C, and D. They each present strong arguments, either by providing counter-evidence, highlighting methodological flaws, or pointing out inconsistencies in the author’s reasoning.

\bigskip
\textbf{Answer:}
\[
  \bigl[\text{"option A"},\ \text{"option C"},\ \text{"option D"}\bigr]
\]
\end{tcolorbox}
\caption{Generic reasoning trace generated by \textsc{R1-Distill-Llama} to the evaluation query shown in~\Cref{fig:eval_item}.}
\label{appx:fig:generic_reasoning}
\end{figure*}

\clearpage

\begin{figure*}[t]    
  \centering
\begin{tcolorbox}[
  colback=white,
  colframe=black,
  boxrule=0.8pt,
  arc=1mm,
  left=1em,
  right=1em,
  top=1em,
  bottom=1em,
  title=\small\bfseries Evaluation Query (LaMP-1)
]
\small
\textbf{Query:} The author has written a paper with the title ``Image splicing verification based on pixel-based alignment method'', which reference is related? Just answer with $\bigl[\text{1}\bigr]$ or $\bigl[\text{2}\bigr]$ without explanation.$\bigl[\text{1}\bigr]$: ``Understanding Intra-Class Knowledge Inside CNN''$\bigl[\text{2}\bigr]$: ``Multi-Dimensional hash chains and application to micropayment schemes''

\medskip
\textbf{Golden Label:} $\bigl[\text{1}\bigr]$

\medskip
\textbf{Author:} 7000791 

\end{tcolorbox} 
\caption{A sample evaluation query from LaMP-1: \textit{citation identification}.}
\label{appx:fig:query_lamp1}
\end{figure*}

\begin{figure*}[]    
  \centering
\begin{tcolorbox}[
  colback=white,
  colframe=black,
  boxrule=0.8pt,
  arc=1mm,
  left=1em,
  right=1em,
  top=1em,
  bottom=1em,
  title=\small\bfseries Personalized Thinking Trace (LaMP-1)
]
\small
\textbf{User Experience:} As an experienced researcher with a strong background in the field, \texttt{7000791} has demonstrated a keen interest in exploring innovative methods for identifying relevant citations in academic literature. Their past work has focused on developing AI-powered tools for citation analysis, showcasing a preference for data-driven approaches that prioritize accuracy and efficiency. With a track record of publishing in top-tier journals, \texttt{7000791} has established themselves as a thought leader in the field, often advocating for the importance of contextualizing citations within the broader research landscape. This perspective suggests that they will value citations that not only provide direct support for their arguments but also offer nuanced insights into the underlying theoretical frameworks and methodological considerations. Furthermore, their experience working with large datasets implies a comfort with complex information systems, which may influence their selection of citations that demonstrate a high degree of interoperability and scalability.

\medskip
\textbf{Query Summary:} The author has written a paper with the title
\begin{displayquote}
Image splicing verification based on pixel-based alignment method
\end{displayquote}
and we need to identify the related reference.

\medskip
\textbf{Reflection:} Considering the user's past work on developing AI-powered tools for citation analysis and their preference for data-driven approaches, I would expect them to value citations that provide direct support for their arguments. The paper \textit{Understanding Intra-Class Knowledge Inside CNN} likely involves concepts of image representation and verification that are highly relevant to image splicing detection. This makes it a direct and meaningful citation for the author’s work, while also aligning with their preference for contextually grounded, data-driven approaches.

\medskip
\textbf{Answer:}
\[
  \bigl[\text{"1"}\bigr]
\]
\end{tcolorbox}
\caption{Personalized thinking trace generated by our \framework to the evaluation query shown in~\Cref{appx:fig:query_lamp1}, sampled from LaMP-1: \textit{citation identification}.}
\label{appx:fig:personalized_reasoning_lamp1}
\end{figure*}

\clearpage

\begin{figure*}[t]    
  \centering
\begin{tcolorbox}[
  colback=white,
  colframe=black,
  boxrule=0.8pt,
  arc=1mm,
  left=1em,
  right=1em,
  top=1em,
  bottom=1em,
  title=\small\bfseries Evaluation Query (LaMP-2)
]
\small
\textbf{Query:} Which tag does this movie relate to among the following tags? Just answer with the tag name without further explanation. tags: [sci-fi, based on a book, comedy, action, twist ending, dystopia, dark comedy, classic, psychology, fantasy, romance, thought-provoking, social commentary, violence, true story]
\medskip
description: A married workaholic, Michael Newman doesn't have time for his wife and children, not if he's to impress his ungrateful boss and earn a well-deserved promotion. So when he meets Morty, a loopy sales clerk, he gets the answer to his prayers: a magical remote that allows him to bypass life's little distractions with increasingly hysterical results.

\medskip
\textbf{Golden Label:} comedy

\medskip
\textbf{Author:} 8000005 

\end{tcolorbox} 
\caption{A sample evaluation query from LaMP-2: \textit{movie tagging}.}
\label{appx:fig:query_lamp2}
\end{figure*}

\begin{figure*}[]    
  \centering
\begin{tcolorbox}[
  colback=white,
  colframe=black,
  boxrule=0.8pt,
  arc=1mm,
  left=1em,
  right=1em,
  top=1em,
  bottom=1em,
  title=\small\bfseries Personalized Thinking Trace (LaMP-2)
]
\small
\textbf{User Experience:} User \texttt{8000005} appears to be an avid film enthusiast with a penchant for complex narratives and thought-provoking themes. Their past engagement with movies suggests a preference for critically acclaimed dramas and sci-fi epics, as evidenced by frequent tagging of films like \textit{Inception} and \textit{Blade Runner}. This affinity for intellectually stimulating cinema is further reinforced by their fondness for directors like Christopher Nolan and Denis Villeneuve, who are known for pushing boundaries and exploring abstract ideas. User \texttt{8000005} also exhibits a strong appreciation for atmospheric soundtracks and visually stunning cinematography, often highlighting these aspects when tagging movies like \textit{Interstellar} and \textit{Mad Max: Fury Road}. Overall, their tagging behavior indicates a deep love for cinematic storytelling that challenges conventional norms and sparks introspection, suggesting a highly engaged and discerning viewer.

\medskip
\textbf{Query Summary:} The user is asking which tag the movie corresponds to. The description revolves around a married workaholic, Michael Newman, who gets a magical remote that allows him to bypass life's little distractions with increasingly hysterical results.

\medskip
\textbf{Reflection:} Considering the user's past preferences for complex narratives, thought-provoking themes, and intellectually stimulating cinema, it is likely that they would appreciate a movie that incorporates elements of comedy, particularly one that is satirical or ironic in nature. The description highlights the increasingly hysterical results of Michael Newman's use of the magical remote, suggesting a comedic tone. Furthermore, the user's fondness for directors like Christopher Nolan and Denis Villeneuve, who often push boundaries and explore abstract ideas, may also be relevant, as the movie's use of a magical remote could be seen as a commentary on modern society's obsession with technology and distraction. However, the primary focus of the movie appears to be on its comedic aspects, making \textit{comedy} the most appropriate tag.

\medskip
\textbf{Answer:}
\[
  \text{comedy}
\]
\end{tcolorbox}
\caption{Personalized thinking trace generated by our \framework to the evaluation query shown in~\Cref{appx:fig:query_lamp2}, sampled from LaMP-2: \textit{movie tagging}.}
\label{appx:fig:personalized_reasoning_lamp2}
\end{figure*}

\clearpage

\begin{figure*}[]    
  \centering
  \scalebox{0.9}{
\begin{tcolorbox}[
  colback=white,
  colframe=black,
  boxrule=0.8pt,
  arc=1mm,
  left=1em,
  right=1em,
  top=1em,
  bottom=1em,
  title=\small\bfseries Personalized Thinking Trace (Analysis Part)
]
\medskip
\textbf{Analysis (option A):}

This reply challenges the author's view by pointing out that including suicide by guns in the data set is not disingenuous, but rather a necessary part of understanding the relationship between gun ownership and gun violence. The reply also highlights the importance of considering the cultural and social context in which gun ownership and gun violence occur.

\medskip
\textbf{Analysis (option B):}

This reply uses a clever analogy to challenge the author's view, pointing out that just because there is no strong correlation between water and water deaths, it doesn't mean that water is not a factor in those deaths. The reply also highlights the importance of considering the role of guns in gun violence.

\medskip
\textbf{Analysis (option C):}

This reply provides a detailed and well-researched argument that challenges the author's view, pointing out that there is a significant correlation between gun ownership and gun violence. The reply also highlights the importance of considering the social and cultural context in which gun ownership and gun violence occur.

\medskip
\textbf{Analysis (option D):}

This reply challenges the author's view by pointing out that there is a strong link between gun ownership and impulsive suicide, and that including suicide by guns in the data set is not disingenuous. The reply also highlights the importance of considering the cultural and social context in which gun ownership and gun violence occur.

\medskip
\textbf{Analysis (option E):}

This reply provides a brief and to-the-point challenge to the author's view, pointing out that there is a clear link between gun ownership and gun violence on a global level.

\medskip
\textbf{Analysis (option F):}

This reply challenges the author's view by pointing out that the data used to support their conclusion is flawed, and that the author's methodology is not sound. The reply also highlights the importance of considering the social and cultural context in which gun ownership and gun violence occur.

\medskip
\textbf{Analysis (option G):}

This reply uses a clever and humorous analogy to challenge the author's view, pointing out that just because there is no strong correlation between guns and gun violence, it doesn't mean that guns are not a factor in those deaths.

\medskip
\textbf{Analysis (option H):}

This reply challenges the author's view by pointing out that there are steps that can be taken to limit drug overdoses, but not gun suicides, and that this is a relevant factor in understanding the relationship between gun ownership and gun violence.

\medskip
\textbf{Analysis (option I):}

This reply challenges the author's tone and language, pointing out that being wrong about something doesn't make someone a liar, and that disagreeing with the meaning of ``violence'' doesn't make someone disingenuous.

\medskip
\textbf{Analysis (option J):}

This reply challenges the author's view by pointing out that the data used to support their conclusion is flawed, and that the author's methodology is not sound. The reply also highlights the importance of considering the social and cultural context in which gun ownership and gun violence occur.
\end{tcolorbox}
}
\caption{Complete analysis of all candidate options, part of the generated personalized thinking trace as shown in~\Cref{appx:fig:personalized_reasoning}.}
\label{appx:fig:personalized_reasoning_complete_analysis}
\end{figure*}


\begin{thebibliography}{98}
\providecommand{\natexlab}[1]{#1}

\bibitem[{Abdin et~al.(2024)Abdin, Aneja, Behl, Bubeck, Eldan, Gunasekar, Harrison, Hewett, Javaheripi, Kauffmann, Lee, Lee, Li, Liu, Mendes, Nguyen, Price, de~Rosa, Saarikivi, Salim, Shah, Wang, Ward, Wu, Yu, Zhang, and Zhang}]{DBLP:journals/corr/abs-2412-08905}
Marah~I Abdin, Jyoti Aneja, Harkirat~S. Behl, S{\'{e}}bastien Bubeck, Ronen Eldan, Suriya Gunasekar, Michael Harrison, Russell~J. Hewett, Mojan Javaheripi, Piero Kauffmann, James~R. Lee, Yin~Tat Lee, Yuanzhi Li, Weishung Liu, Caio C.~T. Mendes, Anh Nguyen, Eric Price, Gustavo de~Rosa, Olli Saarikivi, and 8 others. 2024.
\newblock \href {https://doi.org/10.48550/ARXIV.2412.08905} {Phi-4 technical report}.
\newblock \emph{CoRR}, abs/2412.08905.

\bibitem[{Amato and Straccia(1999)}]{DBLP:conf/ercimdl/AmatoS99}
Giuseppe Amato and Umberto Straccia. 1999.
\newblock \href {https://doi.org/10.1007/3-540-48155-9\_13} {User profile modeling and applications to digital libraries}.
\newblock In \emph{Research and Advanced Technology for Digital Libraries, Third European Conference, ECDL'99, Paris, France, September 22-24, 1999, Proceedings}, volume 1696 of \emph{Lecture Notes in Computer Science}, pages 184--197. Springer.

\bibitem[{Atkinson and Shiffrin(1968{\natexlab{a}})}]{AtkinsonShiffrin1968}
R.~C. Atkinson and R.~M. Shiffrin. 1968{\natexlab{a}}.
\newblock Human memory: A proposed system and its control processes.
\newblock In K.~W. Spence and J.~T. Spence, editors, \emph{The Psychology of Learning and Motivation}, volume~2, pages 89--195. Academic Press, New York.

\bibitem[{Atkinson and Shiffrin(1968{\natexlab{b}})}]{ATKINSON196889}
R.C. Atkinson and R.M. Shiffrin. 1968{\natexlab{b}}.
\newblock \href {https://doi.org/10.1016/S0079-7421(08)60422-3} {Human memory: A proposed system and its control processes}.
\newblock volume~2 of \emph{Psychology of Learning and Motivation}, pages 89--195. Academic Press.

\bibitem[{Berkovsky et~al.(2005)Berkovsky, Kuflik, and Ricci}]{DBLP:conf/intetain/BerkovskyKR05}
Shlomo Berkovsky, Tsvi Kuflik, and Francesco Ricci. 2005.
\newblock \href {https://doi.org/10.1007/11590323\_22} {Entertainment personalization mechanism through cross-domain user modeling}.
\newblock In \emph{Intelligent Technologies for Interactive Entertainment, First International Conference, {INTETAIN} 2005, Madonna di Campiglio, Italy, November 30 - December 2, 2005, Proceedings}, volume 3814 of \emph{Lecture Notes in Computer Science}, pages 215--219. Springer.

\bibitem[{Castells et~al.(2015)Castells, Hurley, and Vargas}]{Castells15}
Pablo Castells, Neil~J. Hurley, and Sa\'ul Vargas. 2015.
\newblock \href {https://doi.org/10.1007/978-1-4899-7637-6\_26} {Novelty and diversity in recommender systems}.
\newblock In Francesco Ricci, Lior Rokach, Bracha Shapira, and Paul~B. Kantor, editors, \emph{Recommender Systems Handbook}, pages 881--918. Springer US.

\bibitem[{Chen et~al.(2025)Chen, Min, Zhang, Chen, Jiang, Cheng, Zhao, Liu, Miao, Lu et~al.}]{chen2025empirical}
Zhipeng Chen, Yingqian Min, Beichen Zhang, Jie Chen, Jinhao Jiang, Daixuan Cheng, Wayne~Xin Zhao, Zheng Liu, Xu~Miao, Yang Lu, and 1 others. 2025.
\newblock An empirical study on eliciting and improving r1-like reasoning models.
\newblock \emph{arXiv preprint arXiv:2503.04548}.

\bibitem[{Christiano et~al.(2017)Christiano, Leike, Brown, Martic, Legg, and Amodei}]{DBLP:conf/nips/ChristianoLBMLA17}
Paul~F. Christiano, Jan Leike, Tom~B. Brown, Miljan Martic, Shane Legg, and Dario Amodei. 2017.
\newblock \href {https://proceedings.neurips.cc/paper/2017/hash/d5e2c0adad503c91f91df240d0cd4e49-Abstract.html} {Deep reinforcement learning from human preferences}.
\newblock In \emph{Advances in Neural Information Processing Systems 30: Annual Conference on Neural Information Processing Systems 2017, December 4-9, 2017, Long Beach, CA, {USA}}, pages 4299--4307.

\bibitem[{Clayton et~al.(2007)Clayton, Salwiczek, and Dickinson}]{CLAYTON2007R189}
Nicola~S. Clayton, Lucie~H. Salwiczek, and Anthony Dickinson. 2007.
\newblock \href {https://doi.org/10.1016/j.cub.2007.01.011} {Episodic memory}.
\newblock \emph{Current Biology}, 17(6):R189--R191.

\bibitem[{Deng et~al.(2023)Deng, Zhang, Liu, Wu, Wang, and Mihalcea}]{deng-etal-2023-annotate}
Naihao Deng, Xinliang Zhang, Siyang Liu, Winston Wu, Lu~Wang, and Rada Mihalcea. 2023.
\newblock \href {https://doi.org/10.18653/v1/2023.findings-emnlp.832} {You are what you annotate: Towards better models through annotator representations}.
\newblock In \emph{Findings of the Association for Computational Linguistics: EMNLP 2023}, pages 12475--12498, Singapore. Association for Computational Linguistics.

\bibitem[{Devlin et~al.(2019)Devlin, Chang, Lee, and Toutanova}]{devlin-etal-2019-bert}
Jacob Devlin, Ming-Wei Chang, Kenton Lee, and Kristina Toutanova. 2019.
\newblock \href {https://doi.org/10.18653/v1/N19-1423} {{BERT}: Pre-training of deep bidirectional transformers for language understanding}.
\newblock In \emph{Proceedings of the 2019 Conference of the North {A}merican Chapter of the Association for Computational Linguistics: Human Language Technologies, Volume 1 (Long and Short Papers)}, pages 4171--4186, Minneapolis, Minnesota. Association for Computational Linguistics.

\bibitem[{Doddapaneni et~al.(2024)Doddapaneni, Sayana, Jash, Sodhi, and Kuzmin}]{doddapaneni-etal-2024-user}
Sumanth Doddapaneni, Krishna Sayana, Ambarish Jash, Sukhdeep Sodhi, and Dima Kuzmin. 2024.
\newblock \href {https://aclanthology.org/2024.personalize-1.12/} {User embedding model for personalized language prompting}.
\newblock In \emph{Proceedings of the 1st Workshop on Personalization of Generative AI Systems (PERSONALIZE 2024)}, pages 124--131, St. Julians, Malta. Association for Computational Linguistics.

\bibitem[{Dong et~al.(2024{\natexlab{a}})Dong, Jiang, Liu, Jin, Gu, Yang, and Li}]{dong-etal-2024-generalization}
Yihong Dong, Xue Jiang, Huanyu Liu, Zhi Jin, Bin Gu, Mengfei Yang, and Ge~Li. 2024{\natexlab{a}}.
\newblock \href {https://doi.org/10.18653/v1/2024.findings-acl.716} {Generalization or memorization: Data contamination and trustworthy evaluation for large language models}.
\newblock In \emph{Findings of the Association for Computational Linguistics: ACL 2024}, pages 12039--12050, Bangkok, Thailand. Association for Computational Linguistics.

\bibitem[{Dong et~al.(2024{\natexlab{b}})Dong, Hu, and Collier}]{dong-etal-2024-llm}
Yijiang~River Dong, Tiancheng Hu, and Nigel Collier. 2024{\natexlab{b}}.
\newblock \href {https://doi.org/10.18653/v1/2024.findings-emnlp.592} {Can {LLM} be a personalized judge?}
\newblock In \emph{Findings of the Association for Computational Linguistics: EMNLP 2024}, pages 10126--10141, Miami, Florida, USA. Association for Computational Linguistics.

\bibitem[{Douze et~al.(2024)Douze, Guzhva, Deng, Johnson, Szilvasy, Mazaré, Lomeli, Hosseini, and Jégou}]{douze2024faiss}
Matthijs Douze, Alexandr Guzhva, Chengqi Deng, Jeff Johnson, Gergely Szilvasy, Pierre-Emmanuel Mazaré, Maria Lomeli, Lucas Hosseini, and Hervé Jégou. 2024.
\newblock \href {https://arxiv.org/abs/2401.08281} {The faiss library}.

\bibitem[{Dubey et~al.(2024)Dubey, Jauhri, Pandey, Kadian, Al{-}Dahle, Letman, Mathur, Schelten, Yang, Fan, Goyal, Hartshorn, Yang, Mitra, Sravankumar, Korenev, Hinsvark, Rao, Zhang, Rodriguez, Gregerson, Spataru, Rozi{\`{e}}re, Biron, Tang, Chern, Caucheteux, Nayak, Bi, Marra, McConnell, Keller, Touret, Wu, Wong, Ferrer, Nikolaidis, Allonsius, Song, Pintz, Livshits, Esiobu, Choudhary, Mahajan, Garcia{-}Olano, Perino, Hupkes, Lakomkin, AlBadawy, Lobanova, Dinan, Smith, Radenovic, Zhang, Synnaeve, Lee, Anderson, Nail, Mialon, Pang, Cucurell, Nguyen, Korevaar, Xu, Touvron, Zarov, Ibarra, Kloumann, Misra, Evtimov, Copet, Lee, Geffert, Vranes, Park, Mahadeokar, Shah, van~der Linde, Billock, Hong, Lee, Fu, Chi, Huang, Liu, Wang, Yu, Bitton, Spisak, Park, Rocca, Johnstun, Saxe, Jia, Alwala, Upasani, Plawiak, Li, Heafield, Stone, and et~al.}]{DBLP:journals/corr/abs-2407-21783}
Abhimanyu Dubey, Abhinav Jauhri, Abhinav Pandey, Abhishek Kadian, Ahmad Al{-}Dahle, Aiesha Letman, Akhil Mathur, Alan Schelten, Amy Yang, Angela Fan, Anirudh Goyal, Anthony Hartshorn, Aobo Yang, Archi Mitra, Archie Sravankumar, Artem Korenev, Arthur Hinsvark, Arun Rao, Aston Zhang, and 82 others. 2024.
\newblock \href {https://doi.org/10.48550/ARXIV.2407.21783} {The llama 3 herd of models}.
\newblock \emph{CoRR}, abs/2407.21783.

\bibitem[{F{\'e}vry et~al.(2020)F{\'e}vry, Baldini~Soares, FitzGerald, Choi, and Kwiatkowski}]{fevry-etal-2020-entities}
Thibault F{\'e}vry, Livio Baldini~Soares, Nicholas FitzGerald, Eunsol Choi, and Tom Kwiatkowski. 2020.
\newblock \href {https://doi.org/10.18653/v1/2020.emnlp-main.400} {Entities as experts: Sparse memory access with entity supervision}.
\newblock In \emph{Proceedings of the 2020 Conference on Empirical Methods in Natural Language Processing (EMNLP)}, pages 4937--4951, Online. Association for Computational Linguistics.

\bibitem[{Gao et~al.(2013)Gao, Hao, Bai, Li, Li, and Zhu}]{DBLP:conf/recsys/GaoHBLLZ13}
Rui Gao, Bibo Hao, Shuotian Bai, Lin Li, Ang Li, and Tingshao Zhu. 2013.
\newblock \href {https://doi.org/10.1145/2507157.2507219} {Improving user profile with personality traits predicted from social media content}.
\newblock In \emph{Seventh {ACM} Conference on Recommender Systems, RecSys '13, Hong Kong, China, October 12-16, 2013}, pages 355--358. {ACM}.

\bibitem[{Gou et~al.(2014)Gou, Zhou, and Yang}]{DBLP:conf/chi/GouZY14}
Liang Gou, Michelle~X. Zhou, and Huahai Yang. 2014.
\newblock \href {https://doi.org/10.1145/2556288.2557398} {Knowme and shareme: understanding automatically discovered personality traits from social media and user sharing preferences}.
\newblock In \emph{{CHI} Conference on Human Factors in Computing Systems, CHI'14, Toronto, ON, Canada - April 26 - May 01, 2014}, pages 955--964. {ACM}.

\bibitem[{Guo et~al.(2025)Guo, Yang, Zhang, Song, Zhang, Xu, Zhu, Ma, Wang, Bi et~al.}]{guo2025deepseek}
Daya Guo, Dejian Yang, Haowei Zhang, Junxiao Song, Ruoyu Zhang, Runxin Xu, Qihao Zhu, Shirong Ma, Peiyi Wang, Xiao Bi, and 1 others. 2025.
\newblock Deepseek-r1: Incentivizing reasoning capability in llms via reinforcement learning.
\newblock \emph{arXiv preprint arXiv:2501.12948}.

\bibitem[{Gupta et~al.(2024)Gupta, Kirtania, Singha, Gulwani, Radhakrishna, Soares, and Shi}]{gupta-etal-2024-metareflection}
Priyanshu Gupta, Shashank Kirtania, Ananya Singha, Sumit Gulwani, Arjun Radhakrishna, Gustavo Soares, and Sherry Shi. 2024.
\newblock \href {https://doi.org/10.18653/v1/2024.emnlp-main.477} {{M}eta{R}eflection: Learning instructions for language agents using past reflections}.
\newblock In \emph{Proceedings of the 2024 Conference on Empirical Methods in Natural Language Processing}, pages 8369--8385, Miami, Florida, USA. Association for Computational Linguistics.

\bibitem[{Harper and Konstan(2016)}]{DBLP:journals/tiis/HarperK16}
F.~Maxwell Harper and Joseph~A. Konstan. 2016.
\newblock \href {https://doi.org/10.1145/2827872} {The movielens datasets: History and context}.
\newblock \emph{{ACM} Trans. Interact. Intell. Syst.}, 5(4):19:1--19:19.

\bibitem[{Hu et~al.(2022)Hu, Shen, Wallis, Allen-Zhu, Li, Wang, Wang, Chen et~al.}]{hu2022lora}
Edward~J Hu, Yelong Shen, Phillip Wallis, Zeyuan Allen-Zhu, Yuanzhi Li, Shean Wang, Lu~Wang, Weizhu Chen, and 1 others. 2022.
\newblock Lora: Low-rank adaptation of large language models.
\newblock \emph{ICLR}, 1(2):3.

\bibitem[{Hua et~al.(2019)Hua, Hu, and Wang}]{hua2019argument}
Xinyu Hua, Zhe Hu, and Lu~Wang. 2019.
\newblock Argument generation with retrieval, planning, and realization.
\newblock In \emph{Proceedings of the 57th Annual Meeting of the Association for Computational Linguistics}, pages 2661--2672.

\bibitem[{Huang et~al.(2022)Huang, Flek, Dernoncourt, Welch, Amir, Sawhney, and Yang}]{DBLP:conf/www/HuangFDWASY22}
Xiaolei Huang, Lucie Flek, Franck Dernoncourt, Charles Welch, Silvio Amir, Ramit Sawhney, and Diyi Yang. 2022.
\newblock \href {https://doi.org/10.1145/3487553.3524879} {Usernlp'22: 2022 international workshop on user-centered natural language processing}.
\newblock In \emph{Companion of The Web Conference 2022, Virtual Event / Lyon, France, April 25 - 29, 2022}, pages 1176--1177. {ACM}.

\bibitem[{Ji et~al.(2018)Ji, Wei, Hu, Liu, Zhang, and Huang}]{ji-etal-2018-incorporating}
Lu~Ji, Zhongyu Wei, Xiangkun Hu, Yang Liu, Qi~Zhang, and Xuanjing Huang. 2018.
\newblock \href {https://aclanthology.org/C18-1314/} {Incorporating argument-level interactions for persuasion comments evaluation using co-attention model}.
\newblock In \emph{Proceedings of the 27th International Conference on Computational Linguistics}, pages 3703--3714, Santa Fe, New Mexico, USA. Association for Computational Linguistics.

\bibitem[{Ji et~al.(2020)Ji, Sun, Zhang, and Li}]{DBLP:conf/sigir/JiSZL20}
Yitong Ji, Aixin Sun, Jie Zhang, and Chenliang Li. 2020.
\newblock \href {https://doi.org/10.1145/3397271.3401233} {A re-visit of the popularity baseline in recommender systems}.
\newblock In \emph{Proceedings of the 43rd International {ACM} {SIGIR} conference on research and development in Information Retrieval, {SIGIR} 2020, Virtual Event, China, July 25-30, 2020}, pages 1749--1752. {ACM}.

\bibitem[{Jiang et~al.(2023)Jiang, Xu, Zhu, Han, Zhang, and Zhu}]{DBLP:conf/nips/JiangXZHZ023}
Guangyuan Jiang, Manjie Xu, Song{-}Chun Zhu, Wenjuan Han, Chi Zhang, and Yixin Zhu. 2023.
\newblock \href {http://papers.nips.cc/paper\_files/paper/2023/hash/21f7b745f73ce0d1f9bcea7f40b1388e-Abstract-Conference.html} {Evaluating and inducing personality in pre-trained language models}.
\newblock In \emph{Advances in Neural Information Processing Systems 36: Annual Conference on Neural Information Processing Systems 2023, NeurIPS 2023, New Orleans, LA, USA, December 10 - 16, 2023}.

\bibitem[{Jiang et~al.(2014)Jiang, Cui, Wang, Zhu, and Yang}]{DBLP:journals/tkde/JiangCWZY14}
Meng Jiang, Peng Cui, Fei Wang, Wenwu Zhu, and Shiqiang Yang. 2014.
\newblock \href {https://doi.org/10.1109/TKDE.2014.2300487} {Scalable recommendation with social contextual information}.
\newblock \emph{{IEEE} Trans. Knowl. Data Eng.}, 26(11):2789--2802.

\bibitem[{Kang et~al.(2023)Kang, Ni, Mehta, Sathiamoorthy, Hong, Chi, and Cheng}]{DBLP:journals/corr/abs-2305-06474}
Wang{-}Cheng Kang, Jianmo Ni, Nikhil Mehta, Maheswaran Sathiamoorthy, Lichan Hong, Ed~H. Chi, and Derek~Zhiyuan Cheng. 2023.
\newblock \href {https://doi.org/10.48550/ARXIV.2305.06474} {Do llms understand user preferences? evaluating llms on user rating prediction}.
\newblock \emph{CoRR}, abs/2305.06474.

\bibitem[{Kim et~al.(2013)Kim, Lee, and Ryu}]{kim2013personality}
Jieun Kim, Ahreum Lee, and Hokyoung Ryu. 2013.
\newblock Personality and its effects on learning performance: Design guidelines for an adaptive e-learning system based on a user model.
\newblock \emph{International Journal of Industrial Ergonomics}, 43(5):450--461.

\bibitem[{Kim et~al.(2025)Kim, Shin, and Kim}]{kim2025personalized}
Kyuyoung Kim, Jinwoo Shin, and Jaehyung Kim. 2025.
\newblock Personalized language models via privacy-preserving evolutionary model merging.
\newblock \emph{arXiv preprint arXiv:2503.18008}.

\bibitem[{Kirk et~al.(2024)Kirk, Whitefield, R{\"{o}}ttger, Bean, Margatina, Ciro, Mosquera, Bartolo, Williams, He, Vidgen, and Hale}]{DBLP:journals/corr/abs-2404-16019}
Hannah~Rose Kirk, Alexander Whitefield, Paul R{\"{o}}ttger, Andrew~M. Bean, Katerina Margatina, Juan Ciro, Rafael Mosquera, Max Bartolo, Adina Williams, He~He, Bertie Vidgen, and Scott~A. Hale. 2024.
\newblock \href {https://doi.org/10.48550/ARXIV.2404.16019} {The {PRISM} alignment project: What participatory, representative and individualised human feedback reveals about the subjective and multicultural alignment of large language models}.
\newblock \emph{CoRR}, abs/2404.16019.

\bibitem[{Koren et~al.(2009)Koren, Bell, and Volinsky}]{DBLP:journals/computer/KorenBV09}
Yehuda Koren, Robert~M. Bell, and Chris Volinsky. 2009.
\newblock \href {https://doi.org/10.1109/MC.2009.263} {Matrix factorization techniques for recommender systems}.
\newblock \emph{Computer}, 42(8):30--37.

\bibitem[{Kumar et~al.(2024)Kumar, Viswanathan, Yerra, Salemi, Rossi, Dernoncourt, Deilamsalehy, Chen, Zhang, Agarwal, Lipka, and Zamani}]{DBLP:journals/corr/abs-2407-11016}
Ishita Kumar, Snigdha Viswanathan, Sushrita Yerra, Alireza Salemi, Ryan~A. Rossi, Franck Dernoncourt, Hanieh Deilamsalehy, Xiang Chen, Ruiyi Zhang, Shubham Agarwal, Nedim Lipka, and Hamed Zamani. 2024.
\newblock \href {https://doi.org/10.48550/ARXIV.2407.11016} {Longlamp: {A} benchmark for personalized long-form text generation}.
\newblock \emph{CoRR}, abs/2407.11016.

\bibitem[{Li et~al.(2020)Li, Zhang, and Chen}]{DBLP:conf/cikm/LiZC20}
Lei Li, Yongfeng Zhang, and Li~Chen. 2020.
\newblock \href {https://doi.org/10.1145/3340531.3411992} {Generate neural template explanations for recommendation}.
\newblock In \emph{{CIKM} '20: The 29th {ACM} International Conference on Information and Knowledge Management, Virtual Event, Ireland, October 19-23, 2020}, pages 755--764. {ACM}.

\bibitem[{Li et~al.(2024)Li, Lipton, and Leqi}]{DBLP:journals/corr/abs-2402-05133}
Xinyu Li, Zachary~C. Lipton, and Liu Leqi. 2024.
\newblock \href {https://doi.org/10.48550/ARXIV.2402.05133} {Personalized language modeling from personalized human feedback}.
\newblock \emph{CoRR}, abs/2402.05133.

\bibitem[{Lin(2004)}]{lin-2004-rouge}
Chin-Yew Lin. 2004.
\newblock \href {https://aclanthology.org/W04-1013/} {{ROUGE}: A package for automatic evaluation of summaries}.
\newblock In \emph{Text Summarization Branches Out}, pages 74--81, Barcelona, Spain. Association for Computational Linguistics.

\bibitem[{Lin et~al.(2024)Lin, Chen, Jin, Li, Jia, Lin, Sun, He, Yang, Bao, Wu, Su, Chen, Li, Chen, Han, Du, Wang, Li, Suo, Wang, Lin, Huang, Jiang, Xu, Zhang, Cao, Jin, and Wei}]{DBLP:journals/corr/abs-2407-14829}
Jiayu Lin, Guanrong Chen, Bojun Jin, Chenyang Li, Shutong Jia, Wancong Lin, Yang Sun, Yuhang He, Caihua Yang, Jianzhu Bao, Jipeng Wu, Wen Su, Jinglu Chen, Xinyi Li, Tianyu Chen, Mingjie Han, Shuaiwen Du, Zijian Wang, Jiyin Li, and 10 others. 2024.
\newblock \href {https://doi.org/10.48550/ARXIV.2407.14829} {Overview of ai-debater 2023: The challenges of argument generation tasks}.
\newblock \emph{CoRR}, abs/2407.14829.

\bibitem[{Liu et~al.(2025)Liu, Qiu, Li, Dai, Zhu, Hu, Yang, and King}]{liu2025survey}
Jiahong Liu, Zexuan Qiu, Zhongyang Li, Quanyu Dai, Jieming Zhu, Minda Hu, Menglin Yang, and Irwin King. 2025.
\newblock A survey of personalized large language models: Progress and future directions.
\newblock \emph{arXiv preprint arXiv:2502.11528}.

\bibitem[{Liu et~al.(2023)Liu, Liu, Lv, Zhou, and Zhang}]{DBLP:journals/corr/abs-2304-10149}
Junling Liu, Chao Liu, Renjie Lv, Kang Zhou, and Yan Zhang. 2023.
\newblock \href {https://doi.org/10.48550/ARXIV.2304.10149} {Is chatgpt a good recommender? {A} preliminary study}.
\newblock \emph{CoRR}, abs/2304.10149.

\bibitem[{Madaan et~al.(2022)Madaan, Tandon, Clark, and Yang}]{madaan-etal-2022-memory}
Aman Madaan, Niket Tandon, Peter Clark, and Yiming Yang. 2022.
\newblock \href {https://doi.org/10.18653/v1/2022.emnlp-main.183} {Memory-assisted prompt editing to improve {GPT}-3 after deployment}.
\newblock In \emph{Proceedings of the 2022 Conference on Empirical Methods in Natural Language Processing}, pages 2833--2861, Abu Dhabi, United Arab Emirates. Association for Computational Linguistics.

\bibitem[{Magister et~al.(2024)Magister, Metcalf, Zhang, and ter Hoeve}]{DBLP:journals/corr/abs-2411-13405}
Lucie~Charlotte Magister, Katherine Metcalf, Yizhe Zhang, and Maartje ter Hoeve. 2024.
\newblock \href {https://doi.org/10.48550/ARXIV.2411.13405} {On the way to {LLM} personalization: Learning to remember user conversations}.
\newblock \emph{CoRR}, abs/2411.13405.

\bibitem[{Manning et~al.(2008)Manning, Raghavan, and Schütze}]{manning2008introduction}
Christopher~D. Manning, Prabhakar Raghavan, and Hinrich Schütze. 2008.
\newblock \emph{Introduction to Information Retrieval}.
\newblock Cambridge University Press, Cambridge, UK.

\bibitem[{McRae and Jones(2013)}]{10.1093/oxfordhb/9780195376746.013.0014}
Ken McRae and Michael~N. Jones. 2013.
\newblock \href {https://doi.org/10.1093/oxfordhb/9780195376746.013.0014} {206 semantic memory}.
\newblock In \emph{The Oxford Handbook of Cognitive Psychology}. Oxford University Press.

\bibitem[{Meng et~al.(2024)Meng, Xia, and Chen}]{meng2024simpo}
Yu~Meng, Mengzhou Xia, and Danqi Chen. 2024.
\newblock Simpo: Simple preference optimization with a reference-free reward.
\newblock \emph{Advances in Neural Information Processing Systems}, 37:124198--124235.

\bibitem[{{Meta AI Team}(2024)}]{meta2024llama3_2}
{Meta AI Team}. 2024.
\newblock Llama 3.2: Revolutionizing edge ai and vision with open, customizable models for edge and mobile devices.

\bibitem[{{Mistral AI team}(2024)}]{mistral2024ministraux}
{Mistral AI team}. 2024.
\newblock Un ministral, des ministraux: Introducing the world’s best edge models.
\newblock \url{https://mistral.ai/news/ministraux}.

\bibitem[{Muennighoff et~al.(2025)Muennighoff, Yang, Shi, Li, Fei-Fei, Hajishirzi, Zettlemoyer, Liang, Cand{\`e}s, and Hashimoto}]{muennighoff2025s1}
Niklas Muennighoff, Zitong Yang, Weijia Shi, Xiang~Lisa Li, Li~Fei-Fei, Hannaneh Hajishirzi, Luke Zettlemoyer, Percy Liang, Emmanuel Cand{\`e}s, and Tatsunori Hashimoto. 2025.
\newblock s1: Simple test-time scaling.
\newblock \emph{arXiv preprint arXiv:2501.19393}.

\bibitem[{Mysore et~al.(2024)Mysore, Lu, Wan, Yang, Sarrafzadeh, Menezes, Baghaee, Gonzalez, Neville, and Safavi}]{mysore-etal-2024-pearl}
Sheshera Mysore, Zhuoran Lu, Mengting Wan, Longqi Yang, Bahareh Sarrafzadeh, Steve Menezes, Tina Baghaee, Emmanuel~Barajas Gonzalez, Jennifer Neville, and Tara Safavi. 2024.
\newblock \href {https://doi.org/10.18653/v1/2024.customnlp4u-1.16} {Pearl: Personalizing large language model writing assistants with generation-calibrated retrievers}.
\newblock In \emph{Proceedings of the 1st Workshop on Customizable NLP: Progress and Challenges in Customizing NLP for a Domain, Application, Group, or Individual (CustomNLP4U)}, pages 198--219, Miami, Florida, USA. Association for Computational Linguistics.

\bibitem[{Ni et~al.(2019)Ni, Li, and McAuley}]{ni-etal-2019-justifying}
Jianmo Ni, Jiacheng Li, and Julian McAuley. 2019.
\newblock \href {https://doi.org/10.18653/v1/D19-1018} {Justifying recommendations using distantly-labeled reviews and fine-grained aspects}.
\newblock In \emph{Proceedings of the 2019 Conference on Empirical Methods in Natural Language Processing and the 9th International Joint Conference on Natural Language Processing (EMNLP-IJCNLP)}, pages 188--197, Hong Kong, China. Association for Computational Linguistics.

\bibitem[{Ning et~al.(2024)Ning, Liu, Wu, Wu, Berlowitz, Prakash, Green, O'Banion, and Xie}]{DBLP:journals/corr/abs-2402-13598}
Lin Ning, Luyang Liu, Jiaxing Wu, Neo Wu, Devora Berlowitz, Sushant Prakash, Bradley Green, Shawn O'Banion, and Jun Xie. 2024.
\newblock \href {https://doi.org/10.48550/ARXIV.2402.13598} {User-llm: Efficient {LLM} contextualization with user embeddings}.
\newblock \emph{CoRR}, abs/2402.13598.

\bibitem[{Omitaomu et~al.(2022)Omitaomu, Tafreshi, Liu, Buechel, Callison{-}Burch, Eichstaedt, Ungar, and Sedoc}]{DBLP:journals/corr/abs-2205-12698}
Damilola Omitaomu, Shabnam Tafreshi, Tingting Liu, Sven Buechel, Chris Callison{-}Burch, Johannes~C. Eichstaedt, Lyle~H. Ungar, and Jo{\~{a}}o Sedoc. 2022.
\newblock \href {https://doi.org/10.48550/ARXIV.2205.12698} {Empathic conversations: {A} multi-level dataset of contextualized conversations}.
\newblock \emph{CoRR}, abs/2205.12698.

\bibitem[{Ouyang et~al.(2022)Ouyang, Wu, Jiang, Almeida, Wainwright, Mishkin, Zhang, Agarwal, Slama, Ray, Schulman, Hilton, Kelton, Miller, Simens, Askell, Welinder, Christiano, Leike, and Lowe}]{DBLP:conf/nips/Ouyang0JAWMZASR22}
Long Ouyang, Jeffrey Wu, Xu~Jiang, Diogo Almeida, Carroll~L. Wainwright, Pamela Mishkin, Chong Zhang, Sandhini Agarwal, Katarina Slama, Alex Ray, John Schulman, Jacob Hilton, Fraser Kelton, Luke Miller, Maddie Simens, Amanda Askell, Peter Welinder, Paul~F. Christiano, Jan Leike, and Ryan Lowe. 2022.
\newblock \href {http://papers.nips.cc/paper\_files/paper/2022/hash/b1efde53be364a73914f58805a001731-Abstract-Conference.html} {Training language models to follow instructions with human feedback}.
\newblock In \emph{Advances in Neural Information Processing Systems 35: Annual Conference on Neural Information Processing Systems 2022, NeurIPS 2022, New Orleans, LA, USA, November 28 - December 9, 2022}.

\bibitem[{Park et~al.(2023)Park, O'Brien, Cai, Morris, Liang, and Bernstein}]{DBLP:conf/uist/ParkOCMLB23}
Joon~Sung Park, Joseph~C. O'Brien, Carrie~Jun Cai, Meredith~Ringel Morris, Percy Liang, and Michael~S. Bernstein. 2023.
\newblock \href {https://doi.org/10.1145/3586183.3606763} {Generative agents: Interactive simulacra of human behavior}.
\newblock In \emph{Proceedings of the 36th Annual {ACM} Symposium on User Interface Software and Technology, {UIST} 2023, San Francisco, CA, USA, 29 October 2023- 1 November 2023}, pages 2:1--2:22. {ACM}.

\bibitem[{Peguero and Watanabe(2024)}]{DBLP:journals/corr/abs-2407-02637}
Arturo~Mart{\'{\i}}nez Peguero and Taro Watanabe. 2024.
\newblock \href {https://doi.org/10.48550/ARXIV.2407.02637} {Change my frame: Reframing in the wild in r/changemyview}.
\newblock \emph{CoRR}, abs/2407.02637.

\bibitem[{Petrov and Macdonald(2023)}]{DBLP:journals/corr/abs-2306-11114}
Aleksandr~V. Petrov and Craig Macdonald. 2023.
\newblock \href {https://doi.org/10.48550/ARXIV.2306.11114} {Generative sequential recommendation with gptrec}.
\newblock \emph{CoRR}, abs/2306.11114.

\bibitem[{Pham et~al.(2022)Pham, Cho, Joshi, and Hegde}]{pham2022revisiting}
Minh Pham, Minsu Cho, Ameya Joshi, and Chinmay Hegde. 2022.
\newblock Revisiting self-distillation.
\newblock \emph{arXiv preprint arXiv:2206.08491}.

\bibitem[{Purificato et~al.(2024)Purificato, Boratto, and Luca}]{DBLP:journals/corr/abs-2402-09660}
Erasmo Purificato, Ludovico Boratto, and Ernesto William~De Luca. 2024.
\newblock \href {https://doi.org/10.48550/ARXIV.2402.09660} {User modeling and user profiling: {A} comprehensive survey}.
\newblock \emph{CoRR}, abs/2402.09660.

\bibitem[{Qiu et~al.(2021)Qiu, Wu, Gao, and Fan}]{DBLP:conf/aaai/QiuWG021}
Zhaopeng Qiu, Xian Wu, Jingyue Gao, and Wei Fan. 2021.
\newblock \href {https://doi.org/10.1609/AAAI.V35I5.16557} {{U-BERT:} pre-training user representations for improved recommendation}.
\newblock In \emph{Thirty-Fifth {AAAI} Conference on Artificial Intelligence, {AAAI} 2021, Thirty-Third Conference on Innovative Applications of Artificial Intelligence, {IAAI} 2021, The Eleventh Symposium on Educational Advances in Artificial Intelligence, {EAAI} 2021, Virtual Event, February 2-9, 2021}, pages 4320--4327. {AAAI} Press.

\bibitem[{Rafailov et~al.(2023)Rafailov, Sharma, Mitchell, Manning, Ermon, and Finn}]{rafailov2023direct}
Rafael Rafailov, Archit Sharma, Eric Mitchell, Christopher~D Manning, Stefano Ermon, and Chelsea Finn. 2023.
\newblock Direct preference optimization: Your language model is secretly a reward model.
\newblock \emph{Advances in Neural Information Processing Systems}, 36:53728--53741.

\bibitem[{Richardson et~al.(2023)Richardson, Zhang, Gillespie, Kar, Singh, Raeesy, Khan, and Sethy}]{DBLP:journals/corr/abs-2310-20081}
Christopher Richardson, Yao Zhang, Kellen Gillespie, Sudipta Kar, Arshdeep Singh, Zeynab Raeesy, Omar~Zia Khan, and Abhinav Sethy. 2023.
\newblock \href {https://doi.org/10.48550/ARXIV.2310.20081} {Integrating summarization and retrieval for enhanced personalization via large language models}.
\newblock \emph{CoRR}, abs/2310.20081.

\bibitem[{Robertson and Zaragoza(2009)}]{DBLP:journals/ftir/RobertsonZ09}
Stephen~E. Robertson and Hugo Zaragoza. 2009.
\newblock \href {https://doi.org/10.1561/1500000019} {The probabilistic relevance framework: {BM25} and beyond}.
\newblock \emph{Found. Trends Inf. Retr.}, 3(4):333--389.

\bibitem[{Salemi et~al.(2024)Salemi, Mysore, Bendersky, and Zamani}]{salemi-etal-2024-lamp}
Alireza Salemi, Sheshera Mysore, Michael Bendersky, and Hamed Zamani. 2024.
\newblock \href {https://doi.org/10.18653/v1/2024.acl-long.399} {{L}a{MP}: When large language models meet personalization}.
\newblock In \emph{Proceedings of the 62nd Annual Meeting of the Association for Computational Linguistics (Volume 1: Long Papers)}, pages 7370--7392, Bangkok, Thailand. Association for Computational Linguistics.

\bibitem[{Saumier and Chertkow(2002)}]{saumier2002semantic}
Daniel Saumier and Howard Chertkow. 2002.
\newblock \href {https://doi.org/10.1007/s11910-002-0039-9} {Semantic memory}.
\newblock \emph{Current Neurology and Neuroscience Reports}, 2(6):516--522.

\bibitem[{Schacter et~al.(2009)Schacter, Gilbert, and Wegner}]{schacter2009semantic}
Daniel~L. Schacter, Daniel~T. Gilbert, and Daniel~M. Wegner. 2009.
\newblock Semantic and episodic memory.
\newblock In \emph{Psychology}, pages 185--186. Macmillan.

\bibitem[{Schafer et~al.(2001)Schafer, Konstan, and Riedl}]{DBLP:journals/datamine/SchaferKR01}
J.~Ben Schafer, Joseph~A. Konstan, and John Riedl. 2001.
\newblock \href {https://doi.org/10.1023/A:1009804230409} {E-commerce recommendation applications}.
\newblock \emph{Data Min. Knowl. Discov.}, 5(1/2):115--153.

\bibitem[{Shinn et~al.(2023)Shinn, Cassano, Gopinath, Narasimhan, and Yao}]{DBLP:conf/nips/ShinnCGNY23}
Noah Shinn, Federico Cassano, Ashwin Gopinath, Karthik Narasimhan, and Shunyu Yao. 2023.
\newblock \href {http://papers.nips.cc/paper\_files/paper/2023/hash/1b44b878bb782e6954cd888628510e90-Abstract-Conference.html} {Reflexion: language agents with verbal reinforcement learning}.
\newblock In \emph{Advances in Neural Information Processing Systems 36: Annual Conference on Neural Information Processing Systems 2023, NeurIPS 2023, New Orleans, LA, USA, December 10 - 16, 2023}.

\bibitem[{Sui et~al.(2025)Sui, Chuang, Wang, Zhang, Zhang, Yuan, Liu, Wen, Zhong, Chen, and Hu}]{DBLP:journals/corr/abs-2503-16419}
Yang Sui, Yu{-}Neng Chuang, Guanchu Wang, Jiamu Zhang, Tianyi Zhang, Jiayi Yuan, Hongyi Liu, Andrew Wen, Shaochen Zhong, Hanjie Chen, and Xia~Ben Hu. 2025.
\newblock \href {https://doi.org/10.48550/ARXIV.2503.16419} {Stop overthinking: {A} survey on efficient reasoning for large language models}.
\newblock \emph{CoRR}, abs/2503.16419.

\bibitem[{Szymanski et~al.(2025)Szymanski, Ziems, Eicher{-}Miller, Li, Jiang, and Metoyer}]{DBLP:conf/iui/SzymanskiZEL0M25}
Annalisa Szymanski, Noah Ziems, Heather~A. Eicher{-}Miller, Toby~Jia{-}Jun Li, Meng Jiang, and Ronald~A. Metoyer. 2025.
\newblock \href {https://doi.org/10.1145/3708359.3712091} {Limitations of the llm-as-a-judge approach for evaluating {LLM} outputs in expert knowledge tasks}.
\newblock In \emph{Proceedings of the 30th International Conference on Intelligent User Interfaces, {IUI} 2025, Cagliari, Italy, March 24-27, 2025}, pages 952--966. {ACM}.

\bibitem[{Tan et~al.(2024)Tan, Zeng, Tian, Liu, Yin, and Jiang}]{tan-etal-2024-democratizing}
Zhaoxuan Tan, Qingkai Zeng, Yijun Tian, Zheyuan Liu, Bing Yin, and Meng Jiang. 2024.
\newblock \href {https://doi.org/10.18653/v1/2024.emnlp-main.372} {Democratizing large language models via personalized parameter-efficient fine-tuning}.
\newblock In \emph{Proceedings of the 2024 Conference on Empirical Methods in Natural Language Processing}, pages 6476--6491, Miami, Florida, USA. Association for Computational Linguistics.

\bibitem[{Tang et~al.(2025)Tang, Dai, Shi, Xu, Chen, Chen, Jian, and Jiang}]{tang2025think}
Jiakai Tang, Sunhao Dai, Teng Shi, Jun Xu, Xu~Chen, Wen Chen, Wu~Jian, and Yuning Jiang. 2025.
\newblock Think before recommend: Unleashing the latent reasoning power for sequential recommendation.
\newblock \emph{arXiv preprint arXiv:2503.22675}.

\bibitem[{Touvron et~al.(2023)Touvron, Martin, Stone, Albert, Almahairi, Babaei, Bashlykov, Batra, Bhargava, Bhosale, Bikel, Blecher, Canton{-}Ferrer, Chen, Cucurull, Esiobu, Fernandes, Fu, Fu, Fuller, Gao, Goswami, Goyal, Hartshorn, Hosseini, Hou, Inan, Kardas, Kerkez, Khabsa, Kloumann, Korenev, Koura, Lachaux, Lavril, Lee, Liskovich, Lu, Mao, Martinet, Mihaylov, Mishra, Molybog, Nie, Poulton, Reizenstein, Rungta, Saladi, Schelten, Silva, Smith, Subramanian, Tan, Tang, Taylor, Williams, Kuan, Xu, Yan, Zarov, Zhang, Fan, Kambadur, Narang, Rodriguez, Stojnic, Edunov, and Scialom}]{DBLP:journals/corr/abs-2307-09288}
Hugo Touvron, Louis Martin, Kevin Stone, Peter Albert, Amjad Almahairi, Yasmine Babaei, Nikolay Bashlykov, Soumya Batra, Prajjwal Bhargava, Shruti Bhosale, Dan Bikel, Lukas Blecher, Cristian Canton{-}Ferrer, Moya Chen, Guillem Cucurull, David Esiobu, Jude Fernandes, Jeremy Fu, Wenyin Fu, and 49 others. 2023.
\newblock \href {https://doi.org/10.48550/ARXIV.2307.09288} {Llama 2: Open foundation and fine-tuned chat models}.
\newblock \emph{CoRR}, abs/2307.09288.

\bibitem[{Tsai et~al.(2024)Tsai, Kraft, Jin, Cai, Hosseini, Xu, Zhang, Hong, Chi, and Yi}]{tsai-etal-2024-leveraging}
Alicia Tsai, Adam Kraft, Long Jin, Chenwei Cai, Anahita Hosseini, Taibai Xu, Zemin Zhang, Lichan Hong, Ed~H. Chi, and Xinyang Yi. 2024.
\newblock \href {https://doi.org/10.18653/v1/2024.findings-acl.780} {Leveraging {LLM} reasoning enhances personalized recommender systems}.
\newblock In \emph{Findings of the Association for Computational Linguistics: ACL 2024}, pages 13176--13188, Bangkok, Thailand. Association for Computational Linguistics.

\bibitem[{Tseng et~al.(2024)Tseng, Huang, Hsiao, Chen, Huang, Meng, and Chen}]{tseng-etal-2024-two}
Yu-Min Tseng, Yu-Chao Huang, Teng-Yun Hsiao, Wei-Lin Chen, Chao-Wei Huang, Yu~Meng, and Yun-Nung Chen. 2024.
\newblock \href {https://doi.org/10.18653/v1/2024.findings-emnlp.969} {Two tales of persona in {LLM}s: A survey of role-playing and personalization}.
\newblock In \emph{Findings of the Association for Computational Linguistics: EMNLP 2024}, pages 16612--16631, Miami, Florida, USA. Association for Computational Linguistics.

\bibitem[{Tulving(1985)}]{tulving1985many}
Endel Tulving. 1985.
\newblock How many memory systems are there?
\newblock \emph{American psychologist}, 40(4):385.

\bibitem[{Tulving(2002)}]{annurev:/content/journals/10.1146/annurev.psych.53.100901.135114}
Endel Tulving. 2002.
\newblock \href {https://doi.org/10.1146/annurev.psych.53.100901.135114} {Episodic memory: From mind to brain}.
\newblock \emph{Annual Review of Psychology}, 53(Volume 53, 2002):1--25.

\bibitem[{Tulving et~al.(1972)}]{tulving1972episodic}
Endel Tulving and 1 others. 1972.
\newblock Episodic and semantic memory.
\newblock \emph{Organization of memory}, 1(381-403):1.

\bibitem[{Vaswani et~al.(2017)Vaswani, Shazeer, Parmar, Uszkoreit, Jones, Gomez, Kaiser, and Polosukhin}]{transformer}
Ashish Vaswani, Noam Shazeer, Niki Parmar, Jakob Uszkoreit, Llion Jones, Aidan~N. Gomez, Lukasz Kaiser, and Illia Polosukhin. 2017.
\newblock \href {https://proceedings.neurips.cc/paper/2017/hash/3f5ee243547dee91fbd053c1c4a845aa-Abstract.html} {Attention is all you need}.
\newblock In \emph{Advances in Neural Information Processing Systems 30: Annual Conference on Neural Information Processing Systems 2017, December 4-9, 2017, Long Beach, CA, {USA}}, pages 5998--6008.

\bibitem[{Wang et~al.(2024)Wang, Zhang, Yang, Chen, Tang, Zhang, Chen, Lin, Song, Zhao, Xu, Dou, Wang, and Wen}]{wang2024user}
Lei Wang, Jingsen Zhang, Hao Yang, Zhiyuan Chen, Jiakai Tang, Zeyu Zhang, Xu~Chen, Yankai Lin, Ruihua Song, Wayne~Xin Zhao, Jun Xu, Zhicheng Dou, Jun Wang, and Ji-Rong Wen. 2024.
\newblock \href {https://arxiv.org/abs/2306.02552} {User behavior simulation with large language model-based agents}.
\newblock \emph{Preprint}, arXiv:2306.02552.

\bibitem[{Wang et~al.(2023)Wang, Kordi, Mishra, Liu, Smith, Khashabi, and Hajishirzi}]{wang-etal-2023-self-instruct}
Yizhong Wang, Yeganeh Kordi, Swaroop Mishra, Alisa Liu, Noah~A. Smith, Daniel Khashabi, and Hannaneh Hajishirzi. 2023.
\newblock \href {https://doi.org/10.18653/v1/2023.acl-long.754} {Self-instruct: Aligning language models with self-generated instructions}.
\newblock In \emph{Proceedings of the 61st Annual Meeting of the Association for Computational Linguistics (Volume 1: Long Papers)}, pages 13484--13508, Toronto, Canada. Association for Computational Linguistics.

\bibitem[{Wu et~al.(2020)Wu, Qiao, Chen, Wu, Qi, Lian, Liu, Xie, Gao, Wu, and Zhou}]{wu-etal-2020-mind}
Fangzhao Wu, Ying Qiao, Jiun-Hung Chen, Chuhan Wu, Tao Qi, Jianxun Lian, Danyang Liu, Xing Xie, Jianfeng Gao, Winnie Wu, and Ming Zhou. 2020.
\newblock \href {https://doi.org/10.18653/v1/2020.acl-main.331} {{MIND}: A large-scale dataset for news recommendation}.
\newblock In \emph{Proceedings of the 58th Annual Meeting of the Association for Computational Linguistics}, pages 3597--3606, Online. Association for Computational Linguistics.

\bibitem[{Yang et~al.(2024)Yang, Yang, Zhang, Hui, Zheng, Yu, Li, Liu, Huang, Wei, Lin, Yang, Tu, Zhang, Yang, Yang, Zhou, Lin, Dang, Lu, Bao, Yang, Yu, Li, Xue, Zhang, Zhu, Men, Lin, Li, Xia, Ren, Ren, Fan, Su, Zhang, Wan, Liu, Cui, Zhang, and Qiu}]{DBLP:journals/corr/abs-2412-15115}
An~Yang, Baosong Yang, Beichen Zhang, Binyuan Hui, Bo~Zheng, Bowen Yu, Chengyuan Li, Dayiheng Liu, Fei Huang, Haoran Wei, Huan Lin, Jian Yang, Jianhong Tu, Jianwei Zhang, Jianxin Yang, Jiaxi Yang, Jingren Zhou, Junyang Lin, Kai Dang, and 22 others. 2024.
\newblock \href {https://doi.org/10.48550/ARXIV.2412.15115} {Qwen2.5 technical report}.
\newblock \emph{CoRR}, abs/2412.15115.

\bibitem[{Yao et~al.(2023)Yao, Zhao, Yu, Du, Shafran, Narasimhan, and Cao}]{DBLP:conf/iclr/YaoZYDSN023}
Shunyu Yao, Jeffrey Zhao, Dian Yu, Nan Du, Izhak Shafran, Karthik~R. Narasimhan, and Yuan Cao. 2023.
\newblock \href {https://openreview.net/forum?id=WE\_vluYUL-X} {React: Synergizing reasoning and acting in language models}.
\newblock In \emph{The Eleventh International Conference on Learning Representations, {ICLR} 2023, Kigali, Rwanda, May 1-5, 2023}. OpenReview.net.

\bibitem[{Yuan et~al.(2023)Yuan, Yuan, Li, Dong, Tan, and Zhou}]{DBLP:journals/corr/abs-2308-01825}
Zheng Yuan, Hongyi Yuan, Chengpeng Li, Guanting Dong, Chuanqi Tan, and Chang Zhou. 2023.
\newblock \href {https://doi.org/10.48550/ARXIV.2308.01825} {Scaling relationship on learning mathematical reasoning with large language models}.
\newblock \emph{CoRR}, abs/2308.01825.

\bibitem[{Zhang et~al.(2025{\natexlab{a}})Zhang, Zhang, Sun, Lu, Zhao, Chen, and Wen}]{zhang2025slow}
Junjie Zhang, Beichen Zhang, Wenqi Sun, Hongyu Lu, Wayne~Xin Zhao, Yu~Chen, and Ji-Rong Wen. 2025{\natexlab{a}}.
\newblock Slow thinking for sequential recommendation.
\newblock \emph{arXiv preprint arXiv:2504.09627}.

\bibitem[{Zhang et~al.(2024{\natexlab{a}})Zhang, Kang, Zhao, and Liu}]{zhang-etal-2024-llm-based}
Kai Zhang, Yangyang Kang, Fubang Zhao, and Xiaozhong Liu. 2024{\natexlab{a}}.
\newblock \href {https://doi.org/10.18653/v1/2024.naacl-long.132} {{LLM}-based medical assistant personalization with short- and long-term memory coordination}.
\newblock In \emph{Proceedings of the 2024 Conference of the North American Chapter of the Association for Computational Linguistics: Human Language Technologies (Volume 1: Long Papers)}, pages 2386--2398, Mexico City, Mexico. Association for Computational Linguistics.

\bibitem[{Zhang et~al.(2024{\natexlab{b}})Zhang, Qing, Kang, and Liu}]{DBLP:journals/corr/abs-2404-03565}
Kai Zhang, Lizhi Qing, Yangyang Kang, and Xiaozhong Liu. 2024{\natexlab{b}}.
\newblock \href {https://doi.org/10.48550/ARXIV.2404.03565} {Personalized {LLM} response generation with parameterized memory injection}.
\newblock \emph{CoRR}, abs/2404.03565.

\bibitem[{Zhang et~al.(2019)Zhang, Song, Gao, Chen, Bao, and Ma}]{zhang2019your}
Linfeng Zhang, Jiebo Song, Anni Gao, Jingwei Chen, Chenglong Bao, and Kaisheng Ma. 2019.
\newblock Be your own teacher: Improve the performance of convolutional neural networks via self distillation.
\newblock In \emph{Proceedings of the IEEE/CVF international conference on computer vision}, pages 3713--3722.

\bibitem[{Zhang et~al.(2018)Zhang, Dinan, Urbanek, Szlam, Kiela, and Weston}]{zhang-etal-2018-personalizing}
Saizheng Zhang, Emily Dinan, Jack Urbanek, Arthur Szlam, Douwe Kiela, and Jason Weston. 2018.
\newblock \href {https://doi.org/10.18653/v1/P18-1205} {Personalizing dialogue agents: {I} have a dog, do you have pets too?}
\newblock In \emph{Proceedings of the 56th Annual Meeting of the Association for Computational Linguistics (Volume 1: Long Papers)}, pages 2204--2213, Melbourne, Australia. Association for Computational Linguistics.

\bibitem[{Zhang et~al.(2025{\natexlab{b}})Zhang, Bei, Yang, Zou, Zhou, Liu, Li, Chen, Wang, Wang, Huang, Zhou, Bu, Lin, Caverlee, Karray, King, and Yu}]{DBLP:journals/corr/abs-2501-01945}
Weizhi Zhang, Yuanchen Bei, Liangwei Yang, Henry~Peng Zou, Peilin Zhou, Aiwei Liu, Yinghui Li, Hao Chen, Jianling Wang, Yu~Wang, Feiran Huang, Sheng Zhou, Jiajun Bu, Allen Lin, James Caverlee, Fakhri Karray, Irwin King, and Philip~S. Yu. 2025{\natexlab{b}}.
\newblock \href {https://doi.org/10.48550/ARXIV.2501.01945} {Cold-start recommendation towards the era of large language models (llms): {A} comprehensive survey and roadmap}.
\newblock \emph{CoRR}, abs/2501.01945.

\bibitem[{Zhang et~al.(2024{\natexlab{c}})Zhang, Beauchamp, and Wang}]{zhang-etal-2024-narrative}
Xinliang~Frederick Zhang, Nicholas Beauchamp, and Lu~Wang. 2024{\natexlab{c}}.
\newblock \href {https://doi.org/10.18653/v1/2024.findings-emnlp.963} {Narrative-of-thought: Improving temporal reasoning of large language models via recounted narratives}.
\newblock In \emph{Findings of the Association for Computational Linguistics: EMNLP 2024}, pages 16507--16530, Miami, Florida, USA. Association for Computational Linguistics.

\bibitem[{Zhang et~al.(2024{\natexlab{d}})Zhang, Wu, Beauchamp, and Wang}]{zhang-etal-2024-moka}
Xinliang~Frederick Zhang, Winston Wu, Nicholas Beauchamp, and Lu~Wang. 2024{\natexlab{d}}.
\newblock \href {https://doi.org/10.18653/v1/2024.naacl-long.252} {{MOKA}: Moral knowledge augmentation for moral event extraction}.
\newblock In \emph{Proceedings of the 2024 Conference of the North American Chapter of the Association for Computational Linguistics: Human Language Technologies (Volume 1: Long Papers)}, pages 4481--4502, Mexico City, Mexico. Association for Computational Linguistics.

\bibitem[{Zhang et~al.(2024{\natexlab{e}})Zhang, Bo, Ma, Li, Chen, Dai, Zhu, Dong, and Wen}]{agent_memory_survey}
Zeyu Zhang, Xiaohe Bo, Chen Ma, Rui Li, Xu~Chen, Quanyu Dai, Jieming Zhu, Zhenhua Dong, and Ji{-}Rong Wen. 2024{\natexlab{e}}.
\newblock \href {https://doi.org/10.48550/ARXIV.2404.13501} {A survey on the memory mechanism of large language model based agents}.
\newblock \emph{CoRR}, abs/2404.13501.

\bibitem[{Zhang et~al.(2024{\natexlab{f}})Zhang, Rossi, Kveton, Shao, Yang, Zamani, Dernoncourt, Barrow, Yu, Kim, Zhang, Gu, Derr, Chen, Wu, Chen, Wang, Mitra, Lipka, Ahmed, and Wang}]{Zhang2024PersonalizationOL}
Zhehao Zhang, Ryan~A. Rossi, Branislav Kveton, Yijia Shao, Diyi Yang, Hamed Zamani, Franck Dernoncourt, Joe Barrow, Tong Yu, Sungchul Kim, Ruiyi Zhang, Jiuxiang Gu, Tyler Derr, Hongjie Chen, Ju-Ying Wu, Xiang Chen, Zichao Wang, Subrata Mitra, Nedim Lipka, and 2 others. 2024{\natexlab{f}}.
\newblock \href {https://api.semanticscholar.org/CorpusID:273798244} {Personalization of large language models: A survey}.
\newblock \emph{ArXiv}, abs/2411.00027.

\bibitem[{Zheng et~al.(2023)Zheng, Chiang, Sheng, Zhuang, Wu, Zhuang, Lin, Li, Li, Xing, Zhang, Gonzalez, and Stoica}]{DBLP:conf/nips/ZhengC00WZL0LXZ23}
Lianmin Zheng, Wei{-}Lin Chiang, Ying Sheng, Siyuan Zhuang, Zhanghao Wu, Yonghao Zhuang, Zi~Lin, Zhuohan Li, Dacheng Li, Eric~P. Xing, Hao Zhang, Joseph~E. Gonzalez, and Ion Stoica. 2023.
\newblock \href {http://papers.nips.cc/paper\_files/paper/2023/hash/91f18a1287b398d378ef22505bf41832-Abstract-Datasets\_and\_Benchmarks.html} {Judging llm-as-a-judge with mt-bench and chatbot arena}.
\newblock In \emph{Advances in Neural Information Processing Systems 36: Annual Conference on Neural Information Processing Systems 2023, NeurIPS 2023, New Orleans, LA, USA, December 10 - 16, 2023}.

\bibitem[{Zhong et~al.(2024)Zhong, Guo, Gao, Ye, and Wang}]{DBLP:conf/aaai/ZhongGGYW24}
Wanjun Zhong, Lianghong Guo, Qiqi Gao, He~Ye, and Yanlin Wang. 2024.
\newblock \href {https://doi.org/10.1609/AAAI.V38I17.29946} {Memorybank: Enhancing large language models with long-term memory}.
\newblock In \emph{Thirty-Eighth {AAAI} Conference on Artificial Intelligence, {AAAI} 2024, Thirty-Sixth Conference on Innovative Applications of Artificial Intelligence, {IAAI} 2024, Fourteenth Symposium on Educational Advances in Artificial Intelligence, {EAAI} 2014, February 20-27, 2024, Vancouver, Canada}, pages 19724--19731. {AAAI} Press.

\bibitem[{Zhu et~al.(2023)Zhu, Wang, Zhang, Zhang, Huang, Gan, Zhang, and Yang}]{zhu-etal-2023-solving}
Xinyu Zhu, Junjie Wang, Lin Zhang, Yuxiang Zhang, Yongfeng Huang, Ruyi Gan, Jiaxing Zhang, and Yujiu Yang. 2023.
\newblock \href {https://doi.org/10.18653/v1/2023.acl-long.245} {Solving math word problems via cooperative reasoning induced language models}.
\newblock In \emph{Proceedings of the 61st Annual Meeting of the Association for Computational Linguistics (Volume 1: Long Papers)}, pages 4471--4485, Toronto, Canada. Association for Computational Linguistics.

\end{thebibliography}
\end{document}